\pdfoutput=1

\documentclass[11pt]{article}

\usepackage[preprint]{acl}

\usepackage{times}
\usepackage{latexsym}
\usepackage{amsmath} 
\usepackage{amssymb}
\usepackage{booktabs}
\usepackage{graphicx}
\usepackage{caption}
\usepackage{subcaption}
\usepackage{colortbl}
\usepackage{placeins}
\usepackage{booktabs}
\usepackage{multirow}
\usepackage{array}

\usepackage{tikz}
\usepackage{pgfplots}
\pgfplotsset{compat=1.17}

\usepackage{adjustbox}
\usepackage{tcolorbox}
\usepackage{stfloats}

\usepackage{chato-notes}

\pgfplotsset{
    customlegend/.style={
        legend columns=6, 
        legend style={
            draw=none, 
            column sep=1.5ex, 
            font=\small 
        },
        legend cell align={left}
    }
}

\pgfplotsset{
    /pgfplots/table/@pretable/.style={
        /pgfplots/every axis legend/.append style={
            /tikz/every even column/.append style={column sep=1.5ex}}
        }
}

\definecolor{bilstm}{HTML}{e63946} 
\definecolor{cnn}{HTML}{457b9d} 
\definecolor{decisiontree}{HTML}{2a9d8f} 
\definecolor{knn}{HTML}{f4a261} 
\definecolor{lightgbm}{HTML}{264653} 
\definecolor{linearsvm}{HTML}{e9c46a} 
\definecolor{logisticregression}{HTML}{9d5a6c} 
\definecolor{mlp}{HTML}{a8dadc} 
\definecolor{naivebayesgaussian}{HTML}{5ffad3} 
\definecolor{nonlinearsvm}{HTML}{ffcbf2} 
\definecolor{randomforest}{HTML}{6a0572} 
\definecolor{xgboost}{HTML}{ff6f61} 

\definecolor{myBlue}{RGB}{159, 216, 251}
\usepackage[T1]{fontenc}

\usepackage[utf8]{inputenc}

\usepackage{microtype}

\usepackage{inconsolata}

\usepackage{graphicx}

\usepackage{enumitem}

\author{
  \normalsize
  \textbf{Dario Di Palma\textsuperscript{1}} \hspace{1em}
  \textbf{Alessandro De Bellis\textsuperscript{1}} \hspace{1em}
  \textbf{Giovanni Servedio\textsuperscript{1,2}} \\
  \normalsize
  \textbf{Vito Walter Anelli\textsuperscript{1}} \hspace{1em}
  \textbf{Fedelucio Narducci\textsuperscript{1}} \hspace{1em}
  \textbf{Tommaso Di Noia\textsuperscript{1}} \\
  \normalsize
  \textsuperscript{1}Politecnico di Bari, Italy \hspace{1em}
  \textsuperscript{2}Sapienza University of Rome, Italy \\
  \normalsize  \href{mailto:dario.dipalma@poliba.it,alessandro.debellis@poliba.it, giovanni.servedio@poliba.it}{name.surname@poliba.it}
}

\hypersetup{
    colorlinks=true,
    linkcolor=black,
    urlcolor=black,
    citecolor=black
}

\usepackage{comment}

\usepackage{fancyhdr} 
\makeatletter
\def\ps@firstpagestyle{\ps@fancy}
\makeatother

\title{LLaMAs Have Feelings Too: Unveiling Sentiment and Emotion Representations in LLaMA Models Through Probing\thanks{
This is the authors’ version of the work. The final, published version will appear in the \textit{Proceedings of the 63rd Annual Meeting of the Association for Computational Linguistics (ACL `25)}.\\ 

This work is licensed under a \href{https://creativecommons.org/licenses/by/4.0/}{Creative Commons Attribution 4.0 International License (CC BY 4.0)}.\\

Please cite the official published version when available.}}

\begin{document}
\maketitle

\begin{abstract}
Large Language Models (LLMs) have rapidly become central to NLP, demonstrating their ability to adapt to various tasks through prompting techniques, including sentiment analysis. However, we still have a limited understanding of how these models capture sentiment-related information. This study probes the hidden layers of Llama models to pinpoint where sentiment features are most represented and to assess how this affects sentiment analysis.

Using probe classifiers, we analyze sentiment encoding across layers and scales, identifying the layers and pooling methods that best capture sentiment signals. Our results show that sentiment information is most concentrated in mid-layers for binary polarity tasks, with detection accuracy increasing up to 14\% over prompting techniques. Additionally, we find that in decoder-only models, the last token is not consistently the most informative for sentiment encoding. Finally, this approach enables sentiment tasks to be performed with memory requirements reduced by an average of 57\%.

These insights contribute to a broader understanding of sentiment in LLMs, suggesting layer-specific probing as an effective approach for sentiment tasks beyond prompting, with potential to enhance model utility and reduce memory requirements.
\end{abstract}

\section{Introduction}
Sentiment analysis (SA), which classifies opinions, emotions, and attitudes in text, is a cornerstone of Natural Language Processing (NLP). SA has enabled the development of several applications, including social media monitoring~\cite{DBLP:conf/emnlp/CamachoColladosRRULABALC22}, customer feedback analysis~\cite{DBLP:conf/ijcnlp/LinXYL17}, and opinion mining~\cite{DBLP:conf/naacl/XiaZWLZHSZ21}. It comprises various tasks, such as polarity detection (classifying text as positive, negative, or neutral), emotion classification, and subjectivity detection~\cite{DBLP:journals/jksucis/KhanBKU14}.

Earlier approaches for the sentiment classification task relied on supervised learning algorithms and extensive feature engineering~\cite{DBLP:conf/semeval/Al-MannaiAWNBM14}, requiring large annotated datasets and domain expertise to model sentiment effectively~\cite{DBLP:journals/air/BordoloiB23}. 

However, the advent of pre-trained models such as GPT-2~\cite{radford2019language} has transformed NLP by enabling downstream tasks through prompting techniques~\cite{DBLP:conf/nips/BrownMRSKDNSSAA20, DBLP:conf/recsys/Palma23}, significantly reducing reliance on extensive labeled data.

While prompting has enhanced the applicability of LLMs for sentiment analysis~\cite{DBLP:conf/www/DengBHBB23a, xing2024designing, DBLP:conf/icacs/AhmedAL24, biancofiore2025conversational}, these techniques have often lacked fine-grained control of semantic features, such as context-dependent sentiments or subtle emotional tones. 
Moreover, the encoding of such features within model representations has not been well understood, limiting efforts to optimize and interpret sentiment analysis performance.

Efforts to interpret neural model representations have advanced significantly over time. 
Building on the linear representations hypothesis~\citep{DBLP:conf/naacl/MikolovYZ13}, which suggests that high-level features can be represented as linear directions, researchers have explored where human-interpretable concepts are encoded within LLMs. 
While previous studies have examined concepts such as truthfulness \citep{DBLP:conf/iclr/BurnsYKS23}, honesty \citep{DBLP:conf/emnlp/AzariaM23}, and factual knowledge \citep{DBLP:conf/nips/0002PVPW23}, sentiment encoding remains relatively underexplored, despite its importance in NLP applications.

This study aims to fill this gap by investigating where sentiment information is encoded within Llama models~\cite{DBLP:journals/corr/abs-2302-13971}. We focus on both binary polarity (positive or negative sentiment) and fine-grained emotion detection (joy, sadness, anger, fear, love, and surprise), identifying the model layers that most effectively capture these sentiment concepts.
Specifically, we trained classifiers to identify subspaces corresponding to specific concepts, such as positive sentiment, within model representations.
We refer to these classifiers as probes~\cite{DBLP:conf/iclr/AlainB17}.

Furthermore, unlike previous studies that rely solely on the last token's representation (as a sentence summary) for probe training~\cite{DBLP:conf/iclr/BurnsYKS23, DBLP:conf/emnlp/AzariaM23, DBLP:conf/nips/0002PVPW23}, we evaluate alternative pooling methods for sentence-level sentiment detection.

Our extensive experiments reveal several key insights into how sentiment and emotion are represented in LLMs. We find that (1) sentiment concepts are most detectable in mid-layer representations, while (2) emotions are more discoverable in early layers; (3) selecting the last token does not consistently yield the best results for probe training; (4) representation quality improves with model size; and (5) probe classifiers generally outperform prompting techniques for accurate sentiment and emotion classification. 

In summary, our work makes the following contributions:
\begin{itemize}[itemsep=0pt, parsep=0pt, topsep=0pt, partopsep=0pt, leftmargin=15pt]
    \item We perform a layer-wise analysis of sentiment encoding in Llama models, identifying the layers that most effectively capture sentiment information.
    \item We evaluate six pooling methods to identify the most effective approach for sentence representation in sentiment analysis.
    \item We investigate the impact of model size on sentiment representation by analyzing Llama-3 models of 1B, 3B, and 8B parameters.
    \item We show that Llama-based classifiers outperform Llama in zero-shot, few-shot, and Chain-of-Thought settings while requiring fewer parameters.
    \item We propose \textsc{SentriLlama}, a task-specific adaptation of Llama that identifies and utilizes the most representative layer for sentiment analysis, replacing the language modeling head with a lightweight classification head to significantly reduce inference costs while maintaining state-of-the-art performance.
\end{itemize}

\section{Related Work}

In this section, we outline the evolution of sentiment analysis, trace the development of probing research in neural models, focusing on LLMs, and conclude by comparing our work to recent studies.

\subsection{Sentiment Analysis Meets LLMs}
Sentiment analysis, the task of classifying and extracting subjective information from text, has evolved from lexicon-based approaches using dictionaries of positive and negative words~\cite{DBLP:conf/acl/HatzivassiloglouM97, DBLP:conf/aaai/Wiebe00} to supervised learning methods employing n-gram models and machine learning~\cite{DBLP:conf/emnlp/PangLV02, DBLP:conf/hicss/ChaovalitZ05}. These early methods, limited by their reliance on handcrafted features, paved the way for feature engineering techniques and eventually the deep learning revolution that reshaped sentiment classification.

To improve feature identification, researchers incorporated advanced techniques such as syntactic dependencies~\cite{DBLP:journals/ccsecis/YaoSLHW10}, part-of-speech tagging~\cite{DBLP:conf/acl/Sogaard11}, and negation handling~\cite{ DBLP:conf/starsem/MoranteB12}. The advent of deep learning further transformed sentiment analysis, with models such as Recurrent Neural Networks (RNNs)~\cite{DBLP:conf/emnlp/SocherPWCMNP13} and Convolutional Neural Networks (CNNs)~\cite{DBLP:conf/emnlp/Kim14}, achieving significant advancements by learning representations directly from raw text.

However, the introduction of Transformer architectures~\cite{DBLP:conf/nips/VaswaniSPUJGKP17} marked a paradigm shift. Models like BERT~\cite{devlin-etal-2019-bert} leverage pretraining on vast datasets to extract contextualized representations, and they have been widely used as encoding backbones for downstream tasks. Conversely, GPT~\cite{radford2019language} employs the Transformer-decoder block and frames language modeling as an autoregressive task. Finally, inspired by the transfer learning paradigm,~\citet{DBLP:journals/jmlr/RaffelSRLNMZLL20} introduced T5, an encoder-decoder model for text-to-text tasks, trained on vast amounts of data to capture general language patterns and fine-tuned for a wide range of specific applications.

The introduction of GPT-3~\cite{DBLP:conf/nips/BrownMRSKDNSSAA20} marked a turning point by introducing few-shot learning and demonstrating the effectiveness of prompt-based techniques, significantly reducing the reliance on large labeled datasets for sentiment analysis and other downstream tasks.

Modern models like GPT-4~\cite{DBLP:journals/corr/abs-2303-08774} and Llama~\cite{DBLP:journals/corr/abs-2302-13971}  achieve even greater flexibility through instruction-following capabilities, excelling in zero-shot settings~\cite{DBLP:conf/emnlp/QinZ0CYY23} and outperforming fine-tuned models in sentiment-related tasks~\cite{hasan-etal-2024-zero}.


Nowadays, the latest models used for sentiment analysis include DeBERTaV3~\cite{DBLP:conf/iclr/HeGC23} and RoBERTa-large~\cite{DBLP:journals/corr/abs-1907-11692}, which have demonstrated excellent performance when fine-tuned. Additionally, models like GPT and Llama have shown effectiveness in this domain, leveraging prompting techniques or instruction fine-tuning~\cite{DBLP:conf/ACMse/StigallKANP24, krugmann2024sentiment, liu2023towards, wei2023larger}, even in complex aspect-based scenarios~\cite{bai-etal-2024-compound, zheng2025reassessing}.

\subsection{Probing and LLMs}
Probe classifiers, or \textit{probes}~\cite{DBLP:conf/iclr/AlainB17}, are tools designed to extract specific properties from the intermediate representations of neural models. In LLMs, probing helps unveil the semantics of their representations by identifying fine-grained features encoded at different layers, allowing researchers to systematically quantify and compare model capabilities.

In the context of NLP, research has evolved from early analyses of static word embeddings like Word2Vec~\cite{DBLP:conf/acl/YaghoobzadehKHA19} and GloVe~\cite{DBLP:conf/blackboxnlp/KlubickaK22} to methods investigating the complex layered knowledge within LLMs~\cite{DBLP:journals/corr/abs-2312-04333, DBLP:conf/nesy/PirozelliJFBC24}.
Prior work has probed LLMs for various fine-grained properties, including linguistic properties~\cite{vulic-etal-2020-probing}, factual knowledge~\cite{petroni-etal-2019-language, wu-etal-2023-plms, DBLP:conf/semweb/BellisANS24}, beliefs~\cite{DBLP:conf/emnlp/AzariaM23}, cross-lingual alignment~\cite{DBLP:conf/acl/WangMP24}, logical reasoning capabilities~\cite{DBLP:conf/nesy/ManigrassoSMB24}, privacy leakages~\cite{NEURIPS2023_420678bb, di2025llms}, toxicity~\cite{wen-etal-2023-unveiling, DBLP:conf/emnlp/RoyH0S23}. These studies approach probing either through a prompt-based method, where the LLM's performance is evaluated using specifically designed prompts, or by applying trained classifiers to analyze the model's intermediate layers.


\subsection{Probing LLMs for Sentiment Analysis}
Various studies have assessed the capabilities of LLMs for sentiment analysis tasks. For instance, \citet{fatouros2023transforming} analyzed the performance of ChatGPT-3.5 in financial sentiment analysis, demonstrating performance that exceeds FinBERT~\cite{DBLP:journals/corr/abs-1908-10063}. Similarly, \citet{DBLP:conf/propor/AraujoMF24} investigated ChatGPT's effectiveness in Portuguese sentiment analysis, highlighting its potential value in dataset annotation. \citet{lyu2024llms} investigates the application of causal inference to sentiment analysis and introduces causal prompts to enhance LLM performance in sentiment prediction tasks. Furthermore, \citet{DBLP:conf/naacl/ZhangDLPB24} conducted a systematic evaluation across various sentiment tasks using ChatGPT and different T5 model sizes. Their findings reveal that while LLMs excel in simple (e.g. binary or trinary) zero-shot sentiment classification tasks, they struggle with complex ones (e.g. aspect-based). 

While previous studies have primarily evaluated LLMs as \textit{text-to-text models}, focusing on zero- and few-shot learning capabilities, our work takes a different approach. We investigate the hidden representations within the intermediate layers of transformer architectures to identify where sentiment concepts are encoded and how these insights can inform the development of more efficient and accurate models.


Similar studies have investigated the representation of semantic concepts within LLMs~\cite{DBLP:conf/cikm/AnelliBBNS22}, such as \citet{DBLP:conf/iclr/BurnsYKS23} on truthfulness, \citet{DBLP:conf/emnlp/AzariaM23} on honesty, \citet{DBLP:conf/emnlp/RoyH0S23} on hate speech, and \citet{DBLP:conf/nips/0002PVPW23} on factual knowledge. However, the investigation of sentiment within the hidden representations 
of these models remains comparatively underexplored.

\section{Methodology}
\label{sec:methodology}
In this section, we detail the models used in the experiments, the datasets, the probe classifiers, and the experimental settings, with a focus on ensuring the reproducibility of our work. Furthermore, we have made all the code publicly available\footnote{\href{https://github.com/sisinflab/LLM-SentimentProber.git}{Sentiment Probing Toolkit}}
to enable systematic and efficient probing of LLMs.
\newline

\noindent\textbf{Sentiment Detection in Hidden Space.}
We build on the concept detection framework proposed by \citet{DBLP:conf/icml/RutteABH24}, adapting it specifically for sentiment analysis. In this setup, we define a sentiment concept \(S\) and use a corresponding sentiment analysis dataset \(\mathcal{D} = \{(x_i, y_i)\}_{i=1}^n\), where \( y_i \) represents the labeled sentiment of sentence \( x_i \). Here, \(\text{rep}_\theta(x_i)\) represents an intermediate representation of \(x_i\), generated from a forward pass through the LLM\(_\theta\). The goal is to extract a collection of these representations \(\{\text{rep}_\theta(x_i)\}_{i=1}^n\). We therefore train a classifier \(C_w\) on these representations to predict the presence of the sentiment concept \(S\) (i.e., positive or negative sentiment within \(x_i\)), effectively predicting \(y_i\).

Implementing sentiment detection involves two fundamental design choices: (1) The selection of the intermediate representation \(\text{rep}_\theta\), which may vary depending on the layers or pooling strategies applied within the LLM. (2) The choice of classifier \(C_w\), which serves to distinguish between different sentiment categories based on these embeddings.

In the following, we outline the fundamental structure of Transformer architecture and highlight common techniques for selecting representations.
\noindent\textbf{Choice of Representation.}\label{subsec:representation} 
Current state-of-the-art LLMs are based on the Transformer architecture~\cite{DBLP:conf/nips/VaswaniSPUJGKP17}, where sequential Transformer blocks generate intermediate hidden representations (\(h\)), each with potentially distinct functionalities. Let \( l \in \mathbb{N} \) denote the \(l\)-th layer, and \( \mathbf{x}^{(l)} \in \mathbb{R}^{T \times d_{\text{emb}}} \) represent the output, where \(T\) is the number of tokens and \(d_{\text{emb}}\) is the hidden dimension. A Transformer refines \( \mathbf{x}^{(l)} \) using multi-head attention (MHA) and a feed-forward network (FFN):
\[
h_{\text{attn}}^{(l)} = \text{MHA} \left( \text{LayerNorm} \left( x^{(l)} \right) \right)
\]
\[
h_{\text{resid}}^{(l)} = h_{\text{attn}}^{(l)} + x^{(l)}
\]
\[
h_{\text{ffn}}^{(l)} = \text{FFN} \left( \text{LayerNorm} \left( h_{\text{resid}}^{(l)} \right) \right)
\]
\[
x^{(l+1)} = h_{\text{ffn}}^{(l)} + h_{\text{resid}}^{(l)}
\]
We exploit the Llama-3 architecture~\cite{DBLP:journals/corr/abs-2407-21783}, leveraging its Grouped Query Attention (GQA) and RMSNorm features to extract hidden representations efficiently.

Previous studies have explored various representations, such as the residual stream (\(\mathbf{x}^{(l+1)}\)) \cite{DBLP:journals/corr/abs-2310-06824, DBLP:conf/iclr/BurnsYKS23, DBLP:journals/corr/abs-2310-01405, DBLP:conf/iclr/GurneeT24}, the normalized residual stream \cite{nostalgebraist2020logitlens}, or attention heads \cite{DBLP:conf/nips/0002PVPW23, arditi2023refusal}. Based on our preliminary experiments showing marginally higher detection accuracy, we adopt the residuals stream (\(\text{rep}_\theta(\mathbf{x}) = \mathbf{x}^{(l+1)}\)).


Instead of using the full prompt representation (\(\mathbf{x}_{\text{rep}} \in \mathbb{R}^{T \times d_{\text{emb}}}\), where \(T\) is the token count), we focus on a subset (\(\mathbf{x}_{\text{rep}} \in \mathbb{R}^{t \times d_{\text{emb}}}\)) with \(t \leq T\). Each token representation (\(\mathbf{x}_{\text{rep}}[i, :] \in \mathbb{R}^{d_{\text{emb}}}\), for \(i = 1, \dots, t\)), is treated as an independent feature. This approach focuses on the parts of the prompt most likely to capture the sentiment concept. Prior work has carefully selected a single token~\cite{arditi2023refusal, DBLP:journals/corr/abs-2310-01405, DBLP:conf/iclr/GurneeT24} or relied on the last token of the prompt~\cite{rimsky2023activationsteering, DBLP:journals/corr/abs-2312-01037, DBLP:journals/corr/abs-2310-06824, DBLP:conf/nips/0002PVPW23, DBLP:conf/iclr/BurnsYKS23}. 

In our experiments, we explore six methods for selecting the representations: 
\begin{enumerate}[label=(\arabic*), itemsep=1pt, parsep=1pt, topsep=1pt, partopsep=1pt, leftmargin=18pt]
    \item Mean Pooling (Fig.~\ref{fig:mean_pooling}): Compute the mean activation value across all tokens for each \( d_{\text{embed}} \) dimension, resulting in a single vector where each element corresponds to the average activation of a particular feature (embedding dimension) over the entire sequence.
    \item Last-Token Pooling (Fig.~\ref{fig:lastoken_pooling}): Uses the final token's features of the last token in the sequence.
    \item Max Pooling (Fig.~\ref{fig:max_min_pooling}): Identifies the most prominent feature across all tokens \( T \) for each feature, outputting a vector representing the most dominant features in the sequence.
    \item Min Pooling (Fig.~\ref{fig:max_min_pooling}): Complementary to Max Pooling, producing a vector representing the least dominant features in the sequence.
    \item Concat-Mean-Max-Min Pooling: Concatenates the mean, max, and min pooling, generating a representation of size \( 3 \times d_{\text{embed}} \) that encapsulates multiple aspects of the token embeddings.
    \item Attention Mean Pooling (Fig.~\ref{fig:attention_pooling}): Given a token representation \(\mathbf{x}_{\text{rep}}[i, j]\), where \(i\) denotes the token and \(j\) represents a specific embedding dimension, this pooling method constructs a representation by combining token embeddings with a corresponding importance score. The importance score for each token \(i\) is computed by applying the softmax function to the mean of the token's embedding values. This assigns higher weights to tokens with larger average values. These scores are multiplied element-wise to each dimension of the token embeddings, \(x_{\text{rep}}[i, :]\), to adjust their contribution based on importance. The final pooled representation, \(x_{\text{pooled}}\), is then computed as a weighted sum of these adjusted embeddings, emphasizing the most relevant tokens activations.
    Mathematically:
    \[
    \resizebox{0.43\textwidth}{!}{
    $x_{\text{pooled}} =  
    \sum_{i=1}^{T} \text{softmax}\left(\frac{1}{J} \sum_{j=1}^{J} x_{\text{rep}}[i, j]\right) \cdot x_{\text{rep}}[i, :]$
    }
    \]

\end{enumerate}

A visual representation of the pooling strategies can be found in Appendix \ref{sec:appendix_Pooling}. These approaches offer diverse strategies for selecting representations, enabling a more nuanced understanding of how sentiment information is encoded. The choice of tokens should ideally reveal whether sentiment is distributed across the entire prompt or concentrated in specific tokens. For instance, the sentences ``My name is XYZ and I'm happy'' and ``I'm happy because my name is XYZ'' both convey positive sentiment. However, relying solely on the last token could lead to a suboptimal representation, as it may not adequately capture the sentiment expressed earlier in the sequence.

\noindent\textbf{Choice of Classifier.} With the hidden representations selected, we can train our probing classifier \( C_w \) on the sentiment labels. In our experimental setup we include twelve distinct classifiers, grouped into five categories based on their underlying modeling approach, as summarized in Table~\ref{tab:model_clusters}.

\begin{table}[h!]
\centering
\resizebox{\columnwidth}{!}{%
\begin{tabular}{@{}p{0.35\columnwidth}p{0.6\columnwidth}@{}}
\toprule
\textbf{Type} & \textbf{Models} \\ 
\midrule
\textbf{Linear}       & Logistic Regression, Linear SVM                \\
\cmidrule{2-2}
\textbf{Distance-based} & K-Nearest Neighbors                          \\
\cmidrule{2-2}
\textbf{Tree-based} & Decision Tree, Random Forest, XGBoost, LightGBM  \\
\cmidrule{2-2}
\textbf{Neural Network} & MLP, BiLSTM, CNN                             \\
\cmidrule{2-2}
\textbf{Other} & Non-linear SVM, Gaussian Naive Bayes                  \\
\bottomrule
\end{tabular}%
}
\caption{Clusters of Classifiers Based on Model Type}
\label{tab:model_clusters}
\end{table}

These classifiers were chosen to balance simplicity, interpretability, and the capacity to model complex patterns. Linear models, such as Logistic Regression and Linear SVM, serve as baselines for probing tasks due to their simplicity and their ability to 
identifying and leveraging linear relationships.
Non-linear and neural network models, on the other hand, are included for their ability to capture intricate relationships within the hidden space. 
BiLSTM and CNN, in particular, were selected for their proven ability to generate effective hidden representations~\cite{DBLP:journals/apin/GhafoorJCHCC23}.

All models, except for BiLSTM and CNN, are implemented using the scikit-learn library~\cite{DBLP:journals/jmlr/PedregosaVGMTGBPWDVPCBPD11}. While, BiLSTM and CNN were implemented using PyTorch. For each classifier, we employed the Optuna framework~\cite{optuna_2019} for hyperparameter optimization, focusing on key parameters such as regularization strengths, tree depths, and kernel types. Optimization was performed over five trials~\cite{DBLP:conf/recsys/PaparellaPAN23}, balancing computational efficiency with sufficient exploration of the hyperparameter space.

To ensure reproducibility, we configured random seed initialization and enforced deterministic behavior for CUDA operations. Details are provided in Appendix~\ref{appendix:reproducibility}.

\noindent\textbf{Datasets Details.} 
Our experiments utilize three benchmark datasets for sentiment classification:
IMDB~\cite{maas-EtAl:2011:ACL-HLT2011}, SST-2~\cite{DBLP:conf/emnlp/SocherPWCMNP13}, and Rotten Tomatoes~\cite{Pang+Lee:05a}, all of which contain movie reviews for binary polarity tasks (e.g., positive or negative sentiment). Additionally, we extended the setup to a more complex evaluation, conducting fine-grained sentiment classification using the Emotion dataset~\cite{DBLP:conf/emnlp/SaraviaLHWC18}, which categorizes sentiments into six nuanced classes: joy, sadness, anger, fear, love, and surprise. 
We only preprocessed the IMDB and Emotion datasets because their original sizes, 50K for IMDB and 20K for Emotion, made them impractical for the large number of experiments. Details of the preprocessing steps are provided in Appendix~\ref{sec:appendix_preprocessing}. Table~\ref{tab:datasets} provides details on the number of samples and the train-test splits used in the experiments.

\begin{table}[h!]
\centering
\resizebox{\columnwidth}{!}{%
\begin{tabular}{@{}lcccc@{}}
\toprule
\textbf{Dataset} & \textbf{Train} & \textbf{Test} & \textbf{Labels}   & \textbf{Max Len} \\ \midrule
IMDB            & 7000           & 7000          & pos/neg          & 132             \\
SST-2           & 6920           & 1821          & pos/neg          & 56              \\
Rotten Tomatoes & 8530           & 1066          & pos/neg          & 59              \\
Emotion         & 6000           & 2000          & six labels      & 64              \\ \bottomrule
\end{tabular}%
}
\caption{Dataset statistics.}
\label{tab:datasets}
\end{table}

\section{Experimental Results}
We conduct our experiments on three Llama models, specifically testing the instruction-tuned variants of Llama 3.2-1B, Llama 3.2-3B, and Llama 3.1-8B, all sourced from Hugging Face\footnote{\url{https://huggingface.co/meta-llama}}. For every probing datasets we train the classifier $C_w$ on the training set, and evaluate its detection performance by measuring the accuracy on the test set.

\begin{figure*}[t]
    \centering
    \begin{tabular}{ccc}
        \begin{subfigure}[t]{0.32\textwidth}
            \centering
            \includegraphics[width=\textwidth]{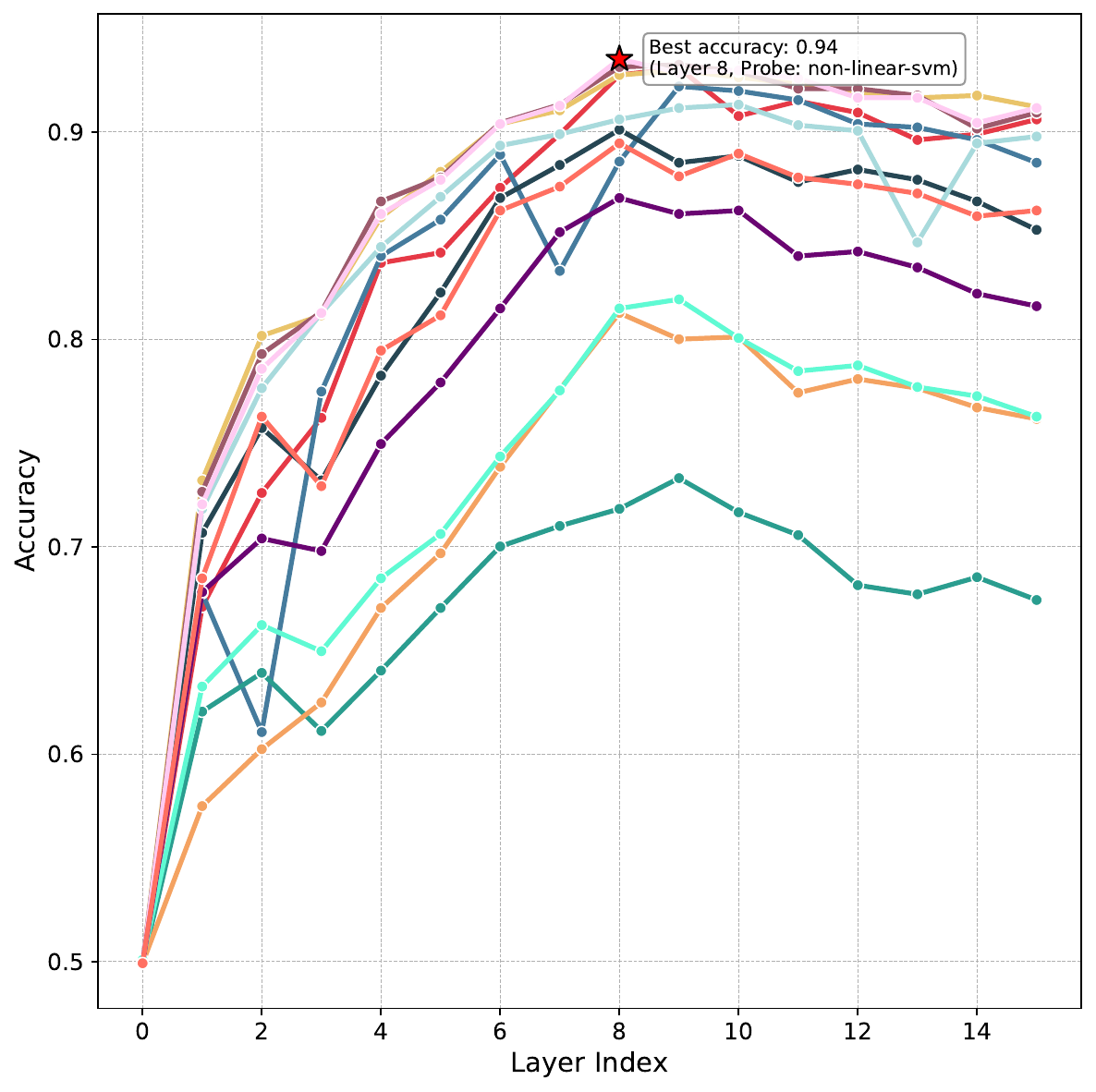}
            \caption{Llama 1B on SST-2}
        \end{subfigure} &
        \begin{subfigure}[t]{0.32\textwidth}
            \centering
            \includegraphics[width=\textwidth]{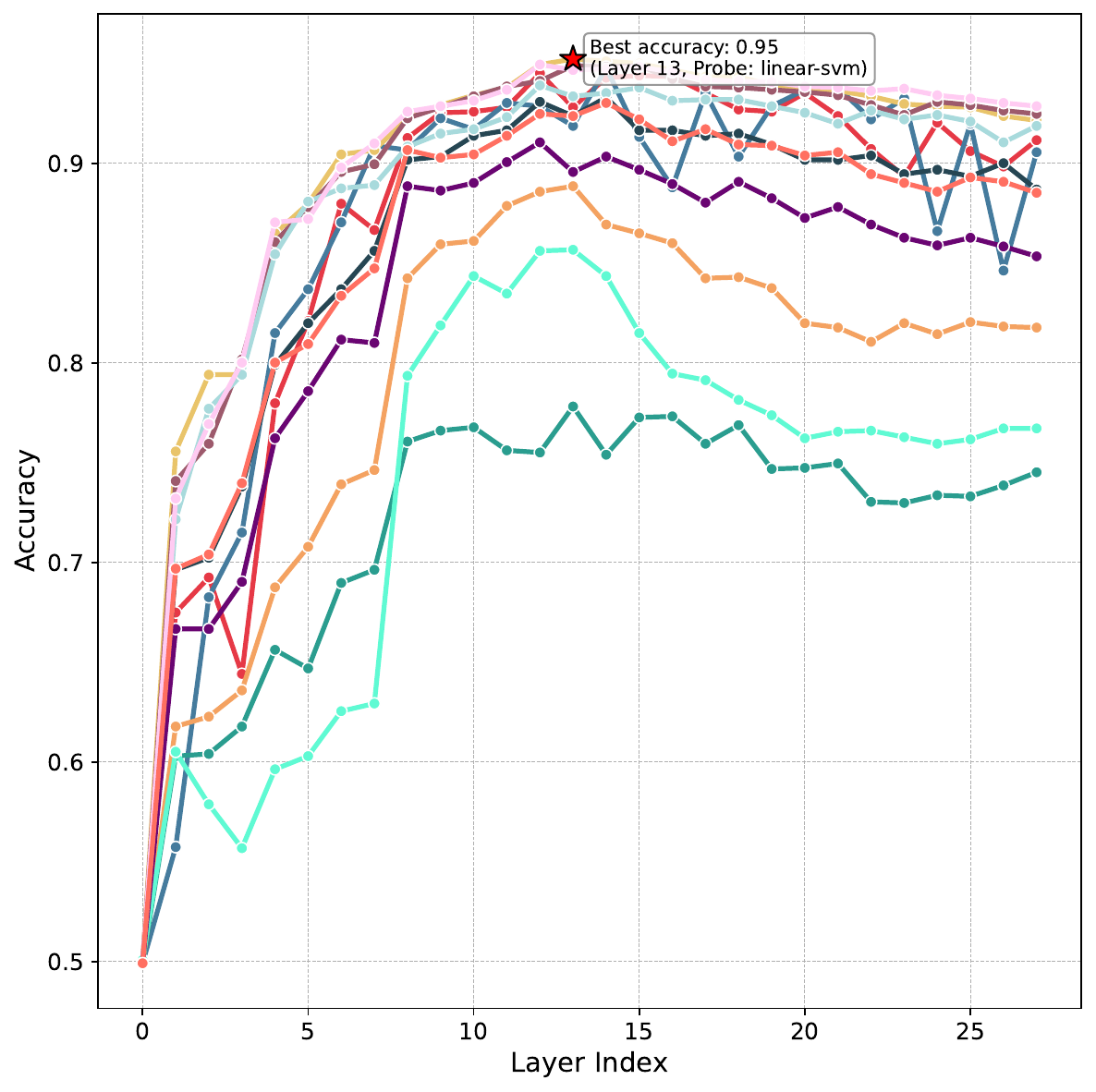}
            \caption{Llama 3B on SST-2}
        \end{subfigure} &
        \begin{subfigure}[t]{0.32\textwidth}
            \centering
            \includegraphics[width=\textwidth]{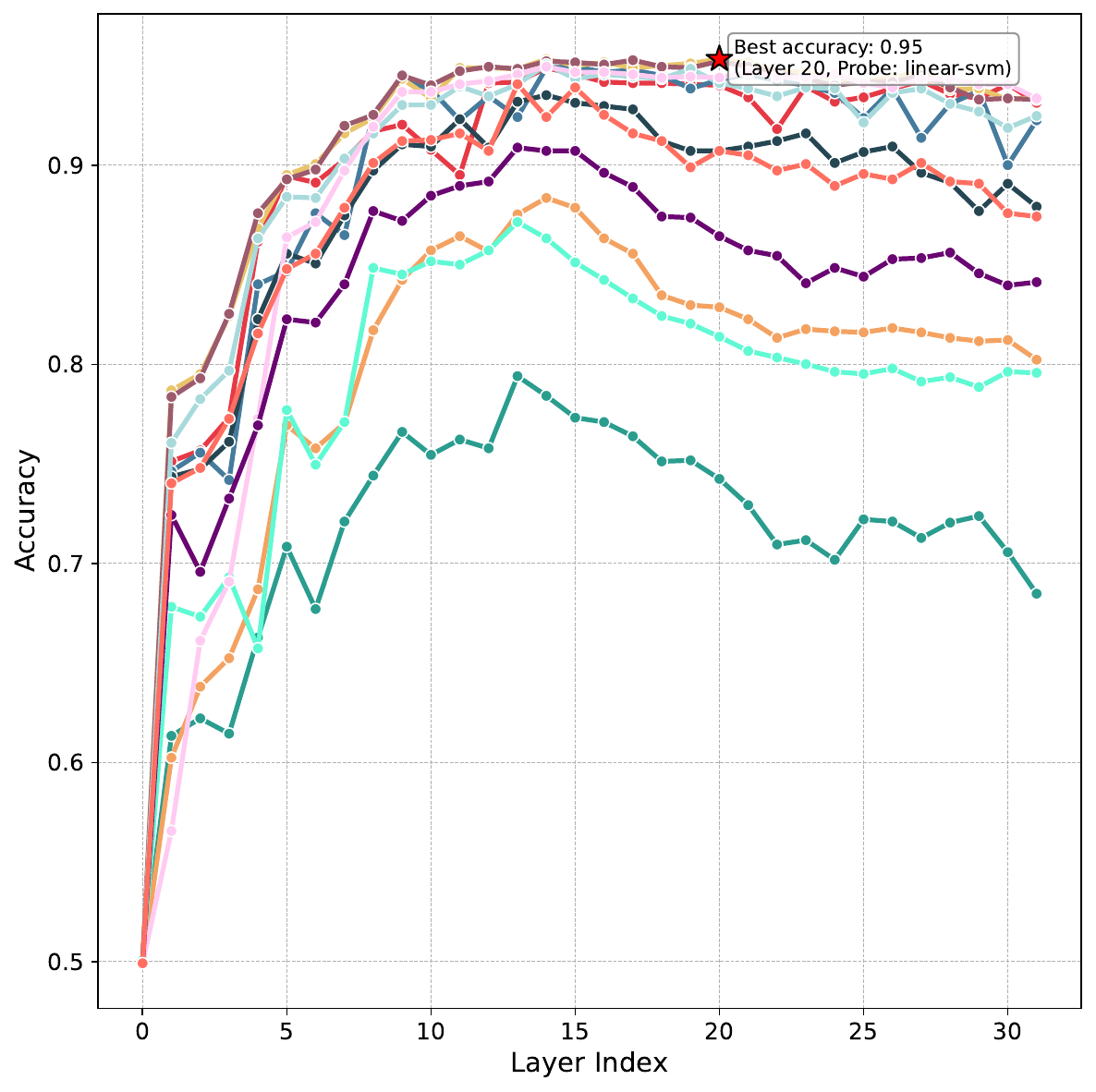}
            \caption{Llama 8B on SST-2}
        \end{subfigure} \\
    \end{tabular}

    \begin{center}
        \resizebox{0.9\textwidth}{!}{ 
            \begin{tikzpicture}
                \begin{axis}[
                    hide axis, 
                    xmin=0, xmax=1, 
                    ymin=0, ymax=1,
                    legend columns=6, 
                    legend style={
                        draw=none, 
                        column sep=1.5ex, 
                        font=\small 
                    },
                    legend entries={
                        bilstm,
                        cnn,
                        decision-tree,
                        knn,
                        lightgbm,
                        linear-svm,
                        logistic-regression,
                        mlp,
                        naive-bayes-gaussian,
                        non-linear-svm,
                        random-forest,
                        xgboost
                    }
                ]
                    \addlegendimage{mark=*, color=bilstm};
                    \addlegendimage{mark=*, color=cnn};
                    \addlegendimage{mark=*, color=decisiontree};
                    \addlegendimage{mark=*, color=knn};
                    \addlegendimage{mark=*, color=lightgbm};
                    \addlegendimage{mark=*, color=linearsvm};
                    \addlegendimage{mark=*, color=logisticregression};
                    \addlegendimage{mark=*, color=mlp};
                    \addlegendimage{mark=*, color=naivebayesgaussian};
                    \addlegendimage{mark=*, color=nonlinearsvm};
                    \addlegendimage{mark=*, color=randomforest};
                    \addlegendimage{mark=*, color=xgboost};
                \end{axis}
            \end{tikzpicture}
        }
    \end{center}
  
    \caption{Layer-wise probing accuracy using the Last-Token approach on SST-2; Appendix~\ref{sec:appendix_Full_LastToken} for additional dataset results.}
    \label{fig:SST2_LastToken_Grid}
\end{figure*}
\begin{table}[ht]
    \centering
    \resizebox{0.5\textwidth}{!}{%
    \begin{tabular}{@{}lccccccc@{}}
    \toprule
    \textbf{Model} & \textbf{Dataset} & \textbf{Layer} & \textbf{Prober} & \textbf{Pooling} & \textbf{Accuracy} \\ 
    \midrule
    \multirow{12}{*}{\shortstack{Llama 3.2 1B \\ (Instruct)}} 
        & \multirow{2}{*}{SST2}
            & \cellcolor{gray!15}10 & \cellcolor{gray!15}Non-linear SVM          & \cellcolor{gray!15}Attn     & \cellcolor{gray!15}0.9450 \\
            & & 10 & Non-linear SVM          & Mean     & 0.9450 \\
            & & 8 & Non-linear SVM          & Last-Token     & 0.9352 \\
        \cmidrule{2-6}
        & \multirow{2}{*}{IMDB}     
            & \cellcolor{gray!15}8   & \cellcolor{gray!15}Logistic Reg.          & \cellcolor{gray!15}Mean    & \cellcolor{gray!15}0.9400 \\
            & & 8   & Logistic Reg.           & Attn    & 0.9396 \\
            & & 8   & Linear SVM           & Last-Token    & 0.9009 \\
        \cmidrule{2-6}
        & \multirow{2}{*}{Rotten}   
            & \cellcolor{gray!15}8   & \cellcolor{gray!15}Linear SVM & \cellcolor{gray!15}Concat       & \cellcolor{gray!15}0.8939 \\
            & & 8   & Logistic Reg. & Concat       & 0.8939 \\
            & & 8   & Non-linear SVM & Last-Token       & 0.8789 \\
        \cmidrule{2-6}
        & \multirow{2}{*}{Emotion}  
            & \cellcolor{gray!15}1  & \cellcolor{gray!15}Linear SVM   & \cellcolor{gray!15}Concat & \cellcolor{gray!15}0.7880 \\
            & & 10  & LightGBM   & Max    & 0.7880 \\
            & & 7  & Linear SVM   & Last-Token    & 0.6830 \\
    \midrule
    \multirow{12}{*}{\shortstack{Llama 3.2 3B \\ (Instruct)}} 
        & \multirow{2}{*}{SST2}     
            &  \cellcolor{gray!15}5  &  \cellcolor{gray!15}Non-linear SVM          &  \cellcolor{gray!15}Concat          &  \cellcolor{gray!15}0.9594 \\
            & & 12  & Non-linear SVM          & Concat          & 0.9577 \\
            & & 13  & Linear SVM          & Last-Token          & 0.9522 \\
        \cmidrule{2-6}
        & \multirow{2}{*}{IMDB}          
            &  \cellcolor{gray!15}4  &  \cellcolor{gray!15}Logistic Reg.         &  \cellcolor{gray!15}Attn     &  \cellcolor{gray!15}0.9523 \\
            & & 14  & Logistic Reg.           & Mean     & 0.9522 \\
            & & 13  & Non-linear SVM          & Last-Token     & 0.9177 \\
        \cmidrule{2-6}
        & \multirow{2}{*}{Rotten}             
            &  \cellcolor{gray!15}3  &  \cellcolor{gray!15}Logistic Reg.     &  \cellcolor{gray!15}Concat   &  \cellcolor{gray!15}0.9090 \\
            & & 3  & CNN     & Mean   & 0.9071 \\
            & & 13  & Linear SVM     & Last-Token   & 0.8949 \\
        \cmidrule{2-6}
        & \multirow{2}{*}{Emotion}  
            &  \cellcolor{gray!15}1   &  \cellcolor{gray!15}MLP          &  \cellcolor{gray!15}Concat          &  \cellcolor{gray!15}0.8220 \\
            & & 0   & LightGBM          & Concat          & 0.8195 \\
            & & 4   & Linear SVM          & Last-Token          & 0.6940 \\
    \midrule
    \multirow{12}{*}{\shortstack{Llama 3.1 8B \\(Instruct)}} 
        & \multirow{2}{*}{SST2}
            &  \cellcolor{gray!15}3  &  \cellcolor{gray!15}Logistic Reg.           &  \cellcolor{gray!15}Concat          &  \cellcolor{gray!15}0.9605 \\
            & & 14  & Logistic Reg.           & Concat          & 0.9599 \\
            & & 14  & Linear SVM           & Last-Token          & 0.9533 \\
        \cmidrule{2-6}
        & \multirow{2}{*}{IMDB}     
            &  \cellcolor{gray!15}4  &  \cellcolor{gray!15}Non-linear SVM          &  \cellcolor{gray!15}Concat   &  \cellcolor{gray!15}0.9579 \\
            & & 16  & Non-linear SVM          & Mean   & 0.9576 \\
            & & 11  & Linear SVM          & Last-Token   & 0.9273 \\
        \cmidrule{2-6}
        & \multirow{2}{*}{Rotten}
            &  \cellcolor{gray!15}9  &  \cellcolor{gray!15}MLP               &  \cellcolor{gray!15}Mean            &  \cellcolor{gray!15}0.9203 \\
            & & 14  & CNN               & Mean            & 0.9203 \\
            & & 14  & Logistic Regression & Last-Token   & 0.9183 \\
        \cmidrule{2-6}
        & \multirow{2}{*}{Emotion}
            &  \cellcolor{gray!15}0   &  \cellcolor{gray!15}Linear SVM   &  \cellcolor{gray!15}Concat    &  \cellcolor{gray!15}0.8685 \\
            & & 0   & LightGBM   & Max    & 0.8655 \\
            & & 4   & Linear SVM   & Last-Token    & 0.6885 \\
    \bottomrule
    \end{tabular}
    }
    \caption{Probing Results Across Different Pooling Methods. See Appendix~\ref{sec:appendix_Confidence_Plot} for confidence level plots across all layers and datasets.}
    \label{tab:pooling_comparation}    
\end{table}

\noindent\textbf{Sentiment detection results.} In our first experiments, we evaluate the detection performance of the different classifier $C_w$. To extract the $\text{rep}_\theta$, following prior work, we focus on the residual stream using the Last-Token approach. We report detection accuracies for each layer in $\text{LLM}_\theta$ and visualize the results in Fig.~\ref{fig:SST2_LastToken_Grid}.

For sentiment detection in binary polarity tasks (i.e., SST-2, IMDB, and Rotten Tomatoes), non-linear SVM, linear SVM, and logistic regression consistently outperform other probing techniques, achieving approximately 90\% accuracy in the middle layers across all model sizes. For fine-grained emotion datasets, linear SVM demonstrates the best performance, reaching around 70\% accuracy in the initial layers across all model sizes. This indicates that Llama models have linear representations for binary sentiment (positive/negative) and fine-grained emotions (joy, sadness, anger, fear, love, and surprise). It is interesting to discover that, although sentiment is not easily discernible during direct interaction with Llama, sentiment and emotion concepts can be linearly detected within certain internal layers.

Furthermore, since prior work often use the last-token representations, we also investigate the effect of different token representation methods. Specifically, we compare the last-token approach with five alternatives: mean, max, min, concatenation, and attention, as previously described. 

The results, summarized in Table~\ref{tab:pooling_comparation}, which shows the top-3 performers by datasets, reveal that concatenating the mean, max, and min of $\text{rep}_\theta$ is often the most effective method for detecting sentiment concepts. Notably, mean and attention pooling also perform strongly, achieving results comparable to concatenation and consistently outperforming last-token pooling. This indicates, that the last-token representation is not always the optimal choice for capturing sentiment concepts.

Furthermore, we discover that combining the max pooling representation with LightGBM yields particularly strong performance. Further investigation reveals that tree-based models, including Random Forest, Decision Tree, and XGBoost, also perform better with max pooling. We hypothesize that the synergy between max pooling and tree-based models arises from max pooling's ability to emphasize dominant features, which, when combined with tree-based models' strength in exploiting high-contrast, threshold-based partitions, leads to an optimal combination. This alignment makes max pooling particularly effective for tree-based models, highlighting how the choice of classifier should also influence the choice of representation, and vice versa.

\section{\textsc{SentriLlama} for Efficient Downstream Tasks}
Sentiment analysis has been a fundamental task since the inception of NLP. Traditionally, there have been two primary approaches to accomplish this task: (1) training a model from scratch or (2) fine-tuning an existing model to meet specific requirements. Recently, a third approach has gained prominence: leveraging well-designed prompts in conjunction with state-of-the-art large language models~\cite{DBLP:conf/www/DengBHBB23a, xing2024designing, DBLP:conf/icacs/AhmedAL24}.

We introduce \textsc{SentriLlama}, a specialized Llama model that leverages layers up to \( L_{\leq i} \) for sentiment tasks, where \( i \) denotes the most representative layer for the task. Based on our earlier analysis, we identified the optimal layer \( L_i \), which retains the most salient features for downstream tasks. By attaching a lightweight, task-specific classification head—such as a LinearSVM—\textsc{SentriLlama} efficiently repurposes the expressive power of the Llama model while discarding unnecessary layers. This approach significantly reduces computational requirements for inference, making it both efficient and task-specific without compromising performance.

Specifically, the total model parameters are calculated as the sum of the input embedding, \( N \)-layer, and LM head parameters. The proposed \textsc{SentriLlama} approach reduces model complexity by retaining only the input embedding and the layers up to the most representative layer (\( i \)), identified in previous experiments, while replacing the LM head with a lightweight classification head, such as a linear SVM. This reduces parameter usage and adapts the architecture for downstream tasks.

To benchmark the effectiveness of \textsc{SentriLlama}, we compare its performance against fine-tuned DeBERTaV3-large\footnote{\href{https://huggingface.co/microsoft/deberta-v3-large}{Microsoft/deberta-v3-large}} and RoBERTa-large\footnote{\href{https://huggingface.co/FacebookAI/roberta-large}{FacebookAI/roberta-large}} across all datasets. Additionally, we evaluate \textsc{SentriLlama} against Llama models under zero-shot, few-shot, and Chain-of-Thought prompting scenarios using carefully designed templates inspired by prior work~\cite{DBLP:conf/www/DengBHBB23a}. Details on the prompt used are provided in Appendix~\ref{sec:appendix_Prompt}. Furthermore, to understand the effect of instruction fine-tuning on the sentiment task, we also include in the comparison the non-instructed version of Llama3.2 (1B). Table~\ref{tab:LLobo_Lama} presents a comparative analysis of the accuracy of \textsc{SentriLlama}, DeBERTa, RoBERTa, and a prompt-based method. The best-performing model is highlighted in \textbf{bold}, while the runner-up is \underline{underlined}.

\begin{table}[htbp]
\centering
\resizebox{0.5\textwidth}{!}{%
\begin{tabular}{@{}lcccc@{}}
\toprule
\textbf{Model}                      & \textbf{SST2} & \textbf{IMDB} & \textbf{Rotten Tomatoes} & \textbf{Emotion} \\ \midrule
\textbf{Instruct-Llama 3.2 (1B)}             &               &               &                          &                  \\
\hspace{2em}Zero-shot               & 0.7210        & 0.6898        & 0.6923                  & 0.2140           \\
\hspace{2em}Few-shot                & 0.6485        & 0.5994        & 0.5994                  & 0.2885           \\ 
\hspace{2em}Chain-of-Thought        & 0.4992        & 0.5000        & 0.5000                  & 0.3475           \\ \midrule
\textbf{Instruct-Llama 3.2 (3B)}             &               &               &                          &                  \\
\hspace{2em}Zero-shot               & 0.7759        & 0.8397        & 0.7279                  & 0.3750           \\
\hspace{2em}Few-shot                & 0.7606        & 0.8528        & 0.7176                  & 0.3045           \\
\hspace{2em}Chain-of-Thought        & 0.9154        & 0.9306        & 0.8743                  & 0.4645           \\ \midrule
\textbf{Instruct-Llama 3.1 (8B)}             &               &               &                          &                  \\
\hspace{2em}Zero-shot               & 0.9341        & 0.9461        & 0.9024                  & 0.4455           \\
\hspace{2em}Few-shot                & 0.9330        & 0.9411        & 0.8968                  & 0.3340           \\
\hspace{2em}Chain-of-Thought        & 0.9165        & 0.9363        & 0.8771                  & 0.5605           \\ \midrule
\textbf{\textsc{SentriLlama} 3.2 (1B)}             & 0.9308        & 0.9445        & 0.8912                  & 0.8015           \\ 
\textbf{\textsc{SentriLlama} 3.2 (1B) Instruct}       & 0.9450        & 0.9400        & 0.8940                  & 0.7880           \\
\textbf{\textsc{SentriLlama} 3.2 (3B) Instruct}       & 0.9594        & 0.9523        & \underline{0.9090}                  & 0.8220           \\ 
\textbf{\textsc{SentriLlama} 3.1 (8B) Instruct}       & \textbf{0.9605} & \textbf{0.9579} & \textbf{0.9203}          & \underline{0.8685}  \\ \midrule
\textbf{DeBERTa V3 Large (418M)}    & \underline{0.9599}        & \underline{0.9534}        & 0.8671                  & \textbf{0.8765}           \\
\textbf{RoBERTa Large (355M)}       & 0.9038        & 0.9430        & 0.8808                  & 0.8416  \\ \bottomrule
\end{tabular}
}
\caption{Comparison of \textsc{SentriLlama} against DeBERTa, RoBERTa, and prompt-based method.}
\label{tab:LLobo_Lama}
\end{table}

For the sentiment downstream task, \textsc{SentriLlama} 8B outperforms all other approaches, achieving approximately 96\% accuracy on SST-2, 96\% on IMDB, 92\% on Rotten Tomatoes, and 87\% on the Emotion dataset. Only DeBERTa surpasses this performance on the Emotion dataset, with an accuracy of 88\%. The prompt-based approach reveals the effectiveness of few-shot and Chain-of-Thought (CoT), achieving comparable results but falling short on the Emotion dataset. These results align with expectations, revealing several insights: (1) increasing the base model size improves performance, (2) the non-instruction version of the 1B model exhibits minimal differences compared to the instructed ones, and (3) DeBERTa and RoBERTa continue to lead the leaderboard in sentiment analysis, outperforming prompt-based methods.
However, it is worth noting that the actual size of \textsc{SentriLlama} differs from the original Llama model. We represent the effective sizes in Table~\ref{tab:LLobo_Lama_Size}.

\begin{table}[htbp]
\centering
\resizebox{0.5\textwidth}{!}{%
\begin{tabular}{@{}lcccc@{}}
\toprule
\textbf{Model}                      & \textbf{SST2} & \textbf{IMDB} & \textbf{Rotten Tomatoes} & \textbf{Emotion} \\ \midrule
\textbf{\textsc{SentriLlama} 3.2 (1B)}             & 810M        & 811M        & 811M                  & 384M           \\ 
\textbf{\textsc{SentriLlama} 3.2 (1B) Instruct}       & 932M        & 810M        & 810M                  & 384M           \\
\textbf{\textsc{SentriLlama} 3.2 (3B) Instruct}       & 2B        & 1.9B        & 1.8B                  & 595M           \\ 
\textbf{\textsc{SentriLlama} 3.1 (8B) Instruct}       & 3.5B & 3.8B & 3.8B          & 743M  \\ \bottomrule
\end{tabular}
}
\caption{Comparison of \textsc{SentriLlama} Sizes Across Datasets.}
\label{tab:LLobo_Lama_Size}
\end{table}
\begin{table}[htbp]
\centering
\resizebox{\columnwidth}{!}{%
\begin{tabular}{@{}lccc@{}}
\toprule
\multirow{2}{*}{\textbf{Model}} & \textbf{Peak GPU} & \textbf{Avg. Time} & \textbf{Throughput} \\ 
& \textbf{Usage} & \textbf{per Sample} & (\textbf{Samples/sec}) \\
\midrule
\textbf{Instruct-Llama 3.2 (1B)} & 2.4 GB & 11.17 ms & 90 \\
\textbf{Instruct-Llama 3.2 (3B)} & 6.2 GB & 18.19 ms & 55 \\
\textbf{Instruct-Llama 3.1 (8B)} & 15.4 GB & 37.73 ms & 48 \\
\midrule
\textbf{\textsc{SentriLlama} 3.2 (1B)} & 1.5 GB & 6.08 ms & 164 \\ 
\textbf{\textsc{SentriLlama} 3.2 (1B) Instruct} & 1.7 GB & 7.98 ms & 125 \\
\textbf{\textsc{SentriLlama} 3.2 (3B) Instruct} & 1.7 GB & 5.09 ms & 196 \\ 
\textbf{\textsc{SentriLlama} 3.1 (8B) Instruct} & 3.2GB & 5.31 ms & 182 \\ 
\midrule
\textbf{DeBERTa V3 Large (418M) } & 845 MB & 22.03 ms & 45 \\
\textbf{RoBERTa Large (355M)} & 692 MB & 8.35 ms & 120 \\ 
\bottomrule
\end{tabular}
}
\caption{Computational efficiency comparison on the SST-2 dataset. See Appendix~\ref{sec:appendix_Comp_Efficiency} for additional dataset results.}
\label{tab:Computational_Efficiency_SST2}
\end{table}

Using \textsc{SentriLlama} reduces the size of the models. For the 1B model (both the standard and instruction-finetuned versions) on SST-2, IMDB, and Rotten Tomatoes, approximately 19\% of the parameters are removed, while for the Emotion dataset, the reduction is up to 61.6\%. For the 3B model on SST-2, IMDB, and Rotten Tomatoes, the reduction is approximately 36.6\%, and for the Emotion dataset, it is 80\%. For the 8B model on SST-2, IMDB, and Rotten Tomatoes, approximately 53.7\% of the parameters are removed, while for the Emotion dataset, the reduction is 90.7\%. This result indicates that the amount of parameters removed depends heavily on the specific task and that it is possible to reuse the highly relevant representations of Llama to improve downstream tasks, going beyond prompting and resulting in a competitive model.

Furthermore, we compare the computational efficiency of \textsc{SentriLlama} with DeBERTa, RoBERTa, and a prompt-based method. In Table~\ref{tab:Computational_Efficiency_SST2}, we present the results for SST-2.

Comparing GPU usage, \textsc{SentriLlama} models demonstrate greater memory efficiency than their Instruct-Llama counterparts. \textsc{SentriLlama} 3.2 (1B) uses 37.5\% less GPU memory than Instruct-Llama 3.2 (1B), while \textsc{SentriLlama} 3.1 (8B) reduces memory usage by 79.2\% compared to Instruct-Llama 3.1 (8B). However, DeBERTa and RoBERTa remain the most memory-efficient, with RoBERTa requiring only 692 MB.

In terms of inference speed, \textsc{SentriLlama} models exhibit a significantly lower average time per sample, indicating they are faster than their Instruct-Llama counterparts. Specifically, \textsc{SentriLlama} 3.2 (1B) is 45.5\% faster than Instruct-Llama 3.2 (1B), \textsc{SentriLlama} 3.2 (3B) is 71.9\% faster than Instruct-Llama 3.2 (3B), and \textsc{SentriLlama} 3.1 (8B) is 85.9\% faster than Instruct-Llama 3.1 (8B). Meanwhile, DeBERTa has a 75.9\% longer inference time compared to \textsc{SentriLlama} 3.2 (8B), while RoBERTa is 36.4\% slower.

Finally, \textsc{SentriLlama} models achieve significantly higher throughput (samples per second) than Instruct-Llama, DeBERTa, and RoBERTa.

Overall, this analysis reveals that \textsc{SentriLlama} models are more efficient in terms of GPU memory usage and inference speed, consistently outperforming Instruct-Llama models across all aspects. Smaller models, such as \textsc{SentriLlama} 3.2 (1B) and \textsc{SentriLlama} 3.2 (3B), offer the best trade-off between memory consumption and speed. Meanwhile, RoBERTa Large provides competitive throughput and latency compared to larger Llama-based models, whereas DeBERTa struggles with slower inference.

\section{Discussion}
Our results show that our method is still far from providing a smaller Llama model that performs as well as similarly sized models like DeBERTa or RoBERTa. The primary limitation stems from the need to retain all preceding layers up to the most representative one. This limitation introduces an element of randomness: if circumstances align favorably and the most representative layer is among the early layers (as observed with the Emotion dataset), the final model's size becomes highly competitive with DeBERTa and RoBERTa. However, if the most representative layer is in the middle layers, the number of parameters—although still lower than the full model—can exceed those of the baseline models.

While we have automated the discovery of the most representative layer and pooling methods (see Table~\ref{tab:pooling_comparation}), the dependence on preceding layers results in a suboptimal outcome. A more robust solution could involve leveraging only the input embeddings and the most representative layer, potentially complemented by layer compression or pruning techniques to reduce computational overhead while preserving performance.

\textit{Is this method generalizable for different downstream tasks beyond sentiment analysis?} While we cannot yet provide a definitive answer on the method's generalizability, prior studies have shown that LLMs capture a broad range of semantic concepts, such as truthfulness, honesty, and factual knowledge \cite{DBLP:conf/iclr/BurnsYKS23, DBLP:conf/emnlp/AzariaM23, DBLP:conf/nips/0002PVPW23}. Our findings suggest that sentiment and emotion are similarly well-represented in these models. We hypothesize that this approach could extend to tasks requiring nuanced linguistic analysis, such as sarcasm detection or intent recognition, assuming the availability of labeled datasets.

However, significant challenges remain. Extending this method to more complex tasks may require additional adaptations. Future research should systematically evaluate this approach across diverse tasks and datasets to uncover its broader potential and limitations.

\section{Conclusion}
In this work, we present a systematic analysis of hidden representations in Llama for sentiment analysis. Our findings demonstrate that for positive/negative sentiment, the middle layers are the most representative across dimensions, whereas for emotion analysis, the most representative layers are found in the initial ones. Additionally, we show that while the current standard for representing a sentence involves using the last token, alternative methods, such as concatenating the max, min, and mean of the representations, yield better results.

Finally, we introduce a novel approach called \textsc{SentriLlama}, which utilizes only a subset of the model's parameters. This approach achieves results comparable to state-of-the-art models and outperforms prompting-based methods. We hope that the introduction of this layer selection approach for Llama will inspire the development of more robust techniques, facilitating the reuse of these large models for downstream tasks and extending their usability beyond text generation.

\section*{Limitations}
This work has two primary limitations. First, the study does not evaluate the proposed approach on domain-specific datasets, such as those from financial, healthcare, or social media domains, nor does it address multilingual sentiment analysis. This gap restricts the generalizability of the findings to a broader range of real-world applications. Second, while the study introduces a method for identifying the most representative layer, there is scope for exploring alternative methodologies within this selection mechanism.



\clearpage
\section*{Appendix}
\appendix
\section{Detailed Preprocessing Workflow}\label{sec:appendix_preprocessing}
The IMDB and Emotion datasets were preprocessed due to their original sizes—50K for IMDB and 20K for Emotion—which made them impractical for the extensive number of experiments. To address this, reduced versions of these datasets were created while preserving their statistical properties. A structured preprocessing workflow was used to prepare the datasets for the experiments, ensuring statistical consistency and adherence to specific constraints, such as sentence length and label distribution. The key steps are outlined below:

\begin{enumerate}[itemsep=1pt, parsep=1pt, topsep=1pt, partopsep=1pt]
    \item \textbf{Initialize Random Seed}: To achieve reproducible datasets, we initialized a random seed (Appendix \ref{appendix:reproducibility}) at the beginning of the workflow, ensuring that the sampling process could be replicated exactly in future runs.
    \item \textbf{Calculate Dataset Statistics}:
        \begin{enumerate}[itemsep=1pt, parsep=1pt, topsep=1pt, partopsep=1pt]
            \item Compute the label distribution to maintain the proportion of samples for each label in the original dataset.
            \item Compute the average sentence length across all samples to preserve similar statistics.
        \end{enumerate}
    \item \textbf{Generate the Dataset}: For each label in the dataset, we followed a sampling process as follows:
        \begin{enumerate}[itemsep=1pt, parsep=1pt, topsep=1pt, partopsep=1pt]
            \item Filter samples by sentence length for the given label.
            \item Random sample to select the required number of samples for each label.
            \item Combine the selected samples for all labels into a single dataset. To ensure randomness, the combined dataset was shuffled before saving.
        \end{enumerate}
    \item \textbf{Validate Reduced Dataset}: The label distribution and average sentence length were recalculated for the reduced dataset. If the average sentence length exceeded the specified constraint, an error was raised, indicating the need for further adjustments.
\end{enumerate}

To provide transparency and facilitate analysis, a summary of the preprocessing steps is reported:
\begin{enumerate}[itemsep=1pt, parsep=1pt, topsep=1pt, partopsep=1pt]
    \item The total number of samples in both the original and reduced datasets.
    \item The distribution of labels before and after reduction.
    \item The average sentence length in both datasets.
\end{enumerate}

This comprehensive preprocessing workflow ensured that the reduced datasets retained the statistical properties of the original datasets while adhering to predefined constraints. The code used to generate the datasets is provided in `$dataset/reduce\_dataset.py$,` along with a detailed PDF report containing all the statistics.

\section{Reproducibility and Determinism}
\label{appendix:reproducibility}

To ensure reproducibility, we consistently set random seeds to 42 across Python’s `random', NumPy, and PyTorch libraries. Furthermore, deterministic behavior was enforced for CUDA-enabled operations by configuring PyTorch’s cuDNN backend and settings environment variables. Further details can be found by examining the $set\_seed()$ function in $utils.py$. This rigorous control of randomness and parallelism ensures consistent experimental results, addressing the stochastic nature of the workflows.

\section{Prompting Approach}\label{sec:appendix_Prompt}

In this section, we illustrate the prompts adopted for the experiments with the Llama models, covering zero-shot, few-shot  and Chain-of-Thought prompting scenarios.

\begin{tcolorbox}[colback=myBlue!5!white, colframe=myBlue!50!black, title=Zero-Shot Prompting for Binary Sentiment]
\small{
\textbf{System:}  "You are an assistant trained to perform strict sentiment and emotion classification.\\
You MUST only respond with the numeric label corresponding to the classification.\\
Do not provide any explanations, reasoning, or any text other than the required numeric value."\\

\textbf{User:} "Classify the sentiment of the following text: `\{text\}'"\\

\textbf{Assistant:} "If the sentiment is positive, respond with `1'. If the sentiment is negative, respond with `0'.\\
No other text, explanation, or formatting."
}
\end{tcolorbox}\label{prompt:zero-shot_binary}

\begin{tcolorbox}[colback=myBlue!5!white, colframe=myBlue!50!black, title=Zero-Shot Prompting for Emotion]
\small{
\textbf{System:} "You are an assistant trained to perform strict sentiment and emotion classification.\\
You MUST only respond with the numeric label corresponding to the classification.\\
Do not provide any explanations, reasoning, or any text other than the required numeric value."\\

\textbf{User:} "Classify the sentiment of the following text: `\{text\}'"\\

\textbf{Assistant:} "Classify the text into one of the following emotions and respond only with the corresponding number:\\
0: Sadness, 1: Joy, 2: Love, 3: Anger, 4: Fear, 5: Surprise.\\
No explanation or additional text."
}
\end{tcolorbox}\label{prompt:zero-shot_emotion}
\begin{tcolorbox}[colback=myBlue!5!white, colframe=myBlue!50!black, title=Few-Shot Prompting for Binary Sentiment]
\small{
\textbf{System:} "You are an assistant trained for sentiment and emotion analysis.\\
You MUST only respond with the correct numeric label.\\
Do not provide explanations or any additional text."\\

\textbf{User:} "Examples:\\
`I love this product!' => 1\\
`I am disappointed with the service.' => 0\\
Classify the following text sentiment:`\{text\}'"\\

\textbf{Assistant:} "If the sentiment is positive, respond with `1'. If the sentiment is negative, respond with `0'.\\
No other text, explanation, or formatting."
}
\end{tcolorbox}\label{prompt:few-shot_binary}

\begin{tcolorbox}[colback=myBlue!5!white, colframe=myBlue!50!black, title=Few-Shot Prompting for Emotion]
\small{
\textbf{System:} "You are an assistant trained for sentiment and emotion analysis.\\
You MUST only respond with the correct numeric label.\\
Do not provide explanations or any additional text."\\

\textbf{User:} "Examples:\\
`This is the worst day of my life.' => 0\\
`I feel so joyful and alive!' => 1\\
`I feel so deeply connected and grateful for you in my life.' => 2\\
`I am so angry right now.' => 3\\
`I’m really scared and worried about what might happen next.' => 4\\
`Wow, I didn't expect that at all! This is completely unexpected!' => 5\\
Classify the following text emotion:`\{text\}'"\\

\textbf{Assistant:} "Respond only with one of the following numbers: 0: Sadness, 1: Joy, 2: Love, 3: Anger, 4: Fear, 5: Surprise.\\
No other text, explanation, or formatting."
}
\end{tcolorbox}\label{prompt:few-shot_Emotion}
\begin{tcolorbox}[colback=myBlue!5!white, colframe=myBlue!50!black, title=Chain-of-Thought Prompting for Binary Sentiment]
\small{
\textbf{System:} "You are an assistant specialized in sentiment and emotion analysis.\\
Think step-by-step through the reasoning process (chain-of-thought) privately, but provide only the final numeric classification as instructed.\\
Do not include reasoning steps in the output."\\

\textbf{User:} "Analyze the sentiment of the following text: `\{text\}' \\
Carefully reason step-by-step to determine the sentiment.\\
Output only `1' for positive sentiment or `0' for negative sentiment as your final response."\\

\textbf{Assistant:} "I will reason step-by-step internally to determine the sentiment.\\
However, my final response will be `1' for positive sentiment or `0' for negative sentiment, with no explanation included in the output."
}
\end{tcolorbox}\label{prompt:Cot_Binary}

\begin{tcolorbox}[colback=myBlue!5!white, colframe=myBlue!50!black, title=Chain-of-Thought Prompting for Emotion]
\small{
\textbf{System:}  "You are an assistant specialized in sentiment and emotion analysis.\\
Think step-by-step through the reasoning process (chain-of-thought) privately, but provide only the final numeric classification as instructed.\\
Do not include reasoning steps in the output."\\

\textbf{User:} "Analyze the emotion of the following text: `\{text\}'\\
Carefully reason step-by-step to identify the best-matching emotion.\\
Output only the corresponding number as your final response: 0: Sadness, 1: Joy, 2: Love, 3: Anger, 4: Fear, 5: Surprise."\\

\textbf{Assistant:} "I will reason step-by-step internally to determine the most appropriate emotion.\\
My final response will be one of the following numbers: 0: Sadness, 1: Joy, 2: Love, 3: Anger, 4: Fear, 5: Surprise.\\ 
No reasoning will be included in the output."
}
\end{tcolorbox}\label{prompt:CoT_Emotion}
\newpage
\section{Pooling Methods}\label{sec:appendix_Pooling}

In this section, we provide a visual clarification of the different pooling strategies adopted in our work. Further details can be found in Section~\ref{sec:methodology}.
\vspace{4em}

\begin{figure}[!ht]
    \centering
    \includegraphics[width=\linewidth]{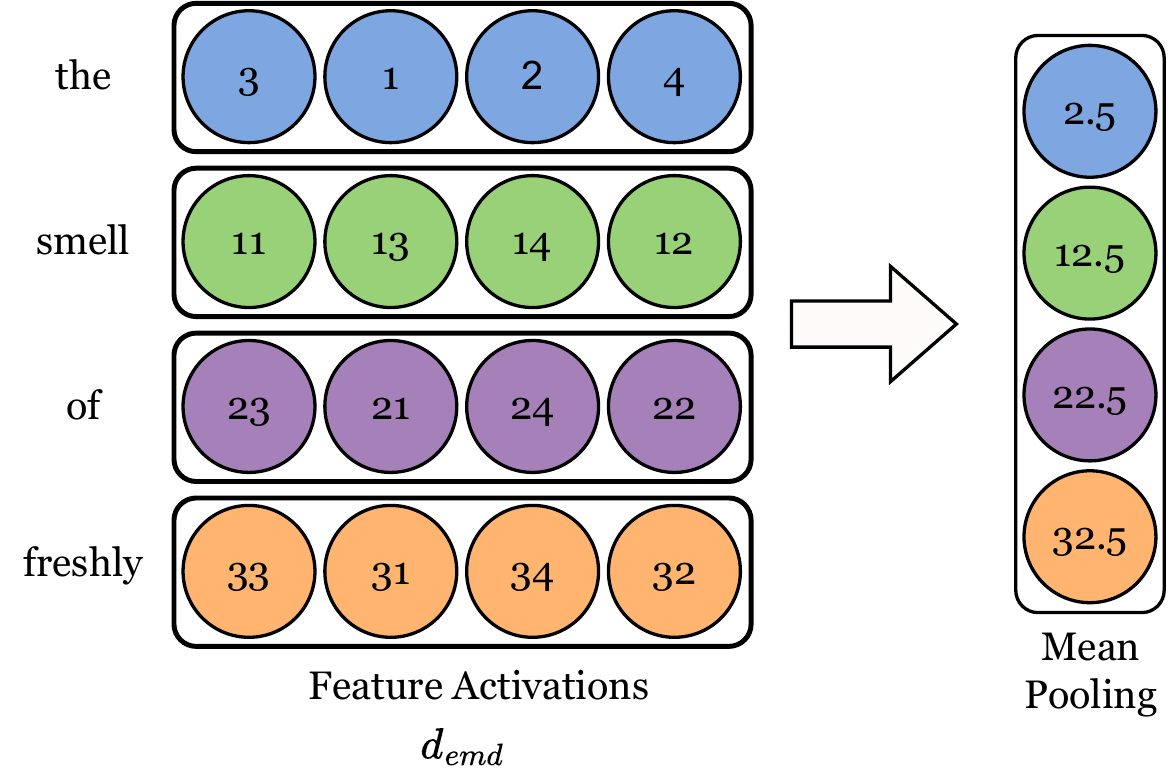}
    \caption{Mean Pooling Visually Explained}
    \label{fig:mean_pooling}
\end{figure}

\begin{figure}[!ht]
    \centering
    \includegraphics[width=\linewidth]{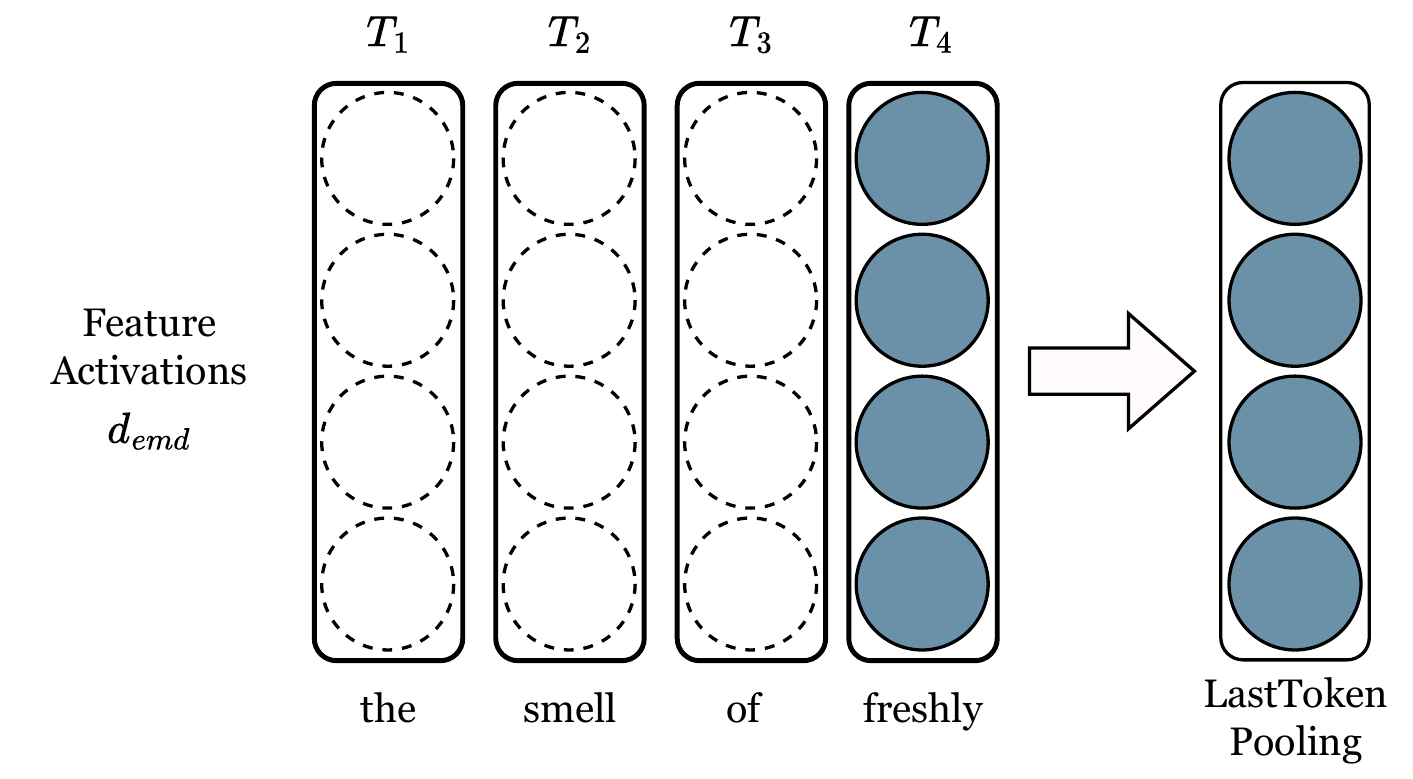}
    \caption{Last Token Pooling Visually Explained}
    \label{fig:lastoken_pooling}
\end{figure}

\begin{figure}[!ht]
    \centering
    \includegraphics[width=\linewidth]{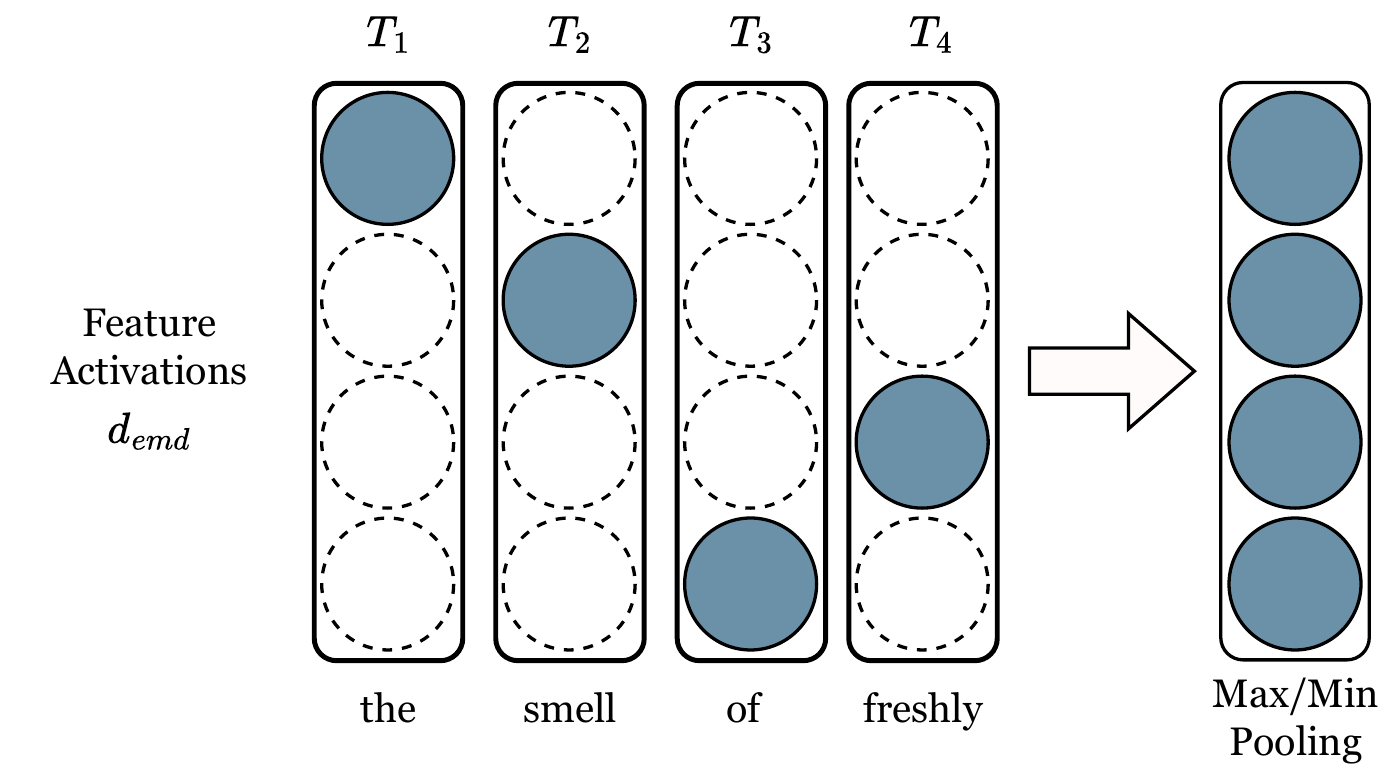}
    \caption{Max and Min Pooling Visually Explained}
    \label{fig:max_min_pooling}
\end{figure}

\begin{figure}[!ht]
    \centering
    \includegraphics[width=\linewidth]{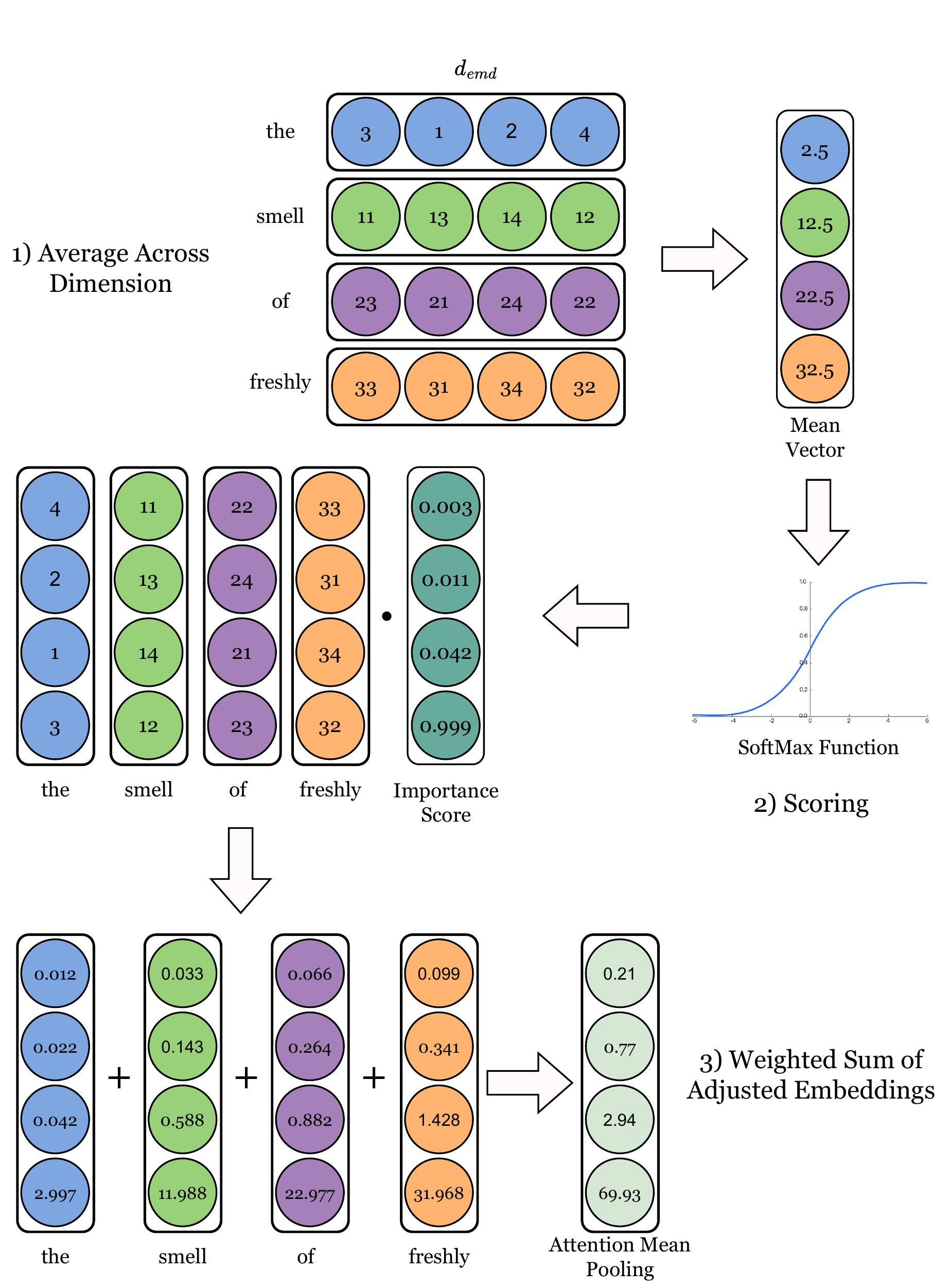}
    \caption{Attention Pooling Visually Explained}
    \label{fig:attention_pooling}
\end{figure}

\onecolumn
\section{Comprehensive Results for the Last-Token Approach} \label{sec:appendix_Full_LastToken}
\begin{figure*}[!htbp]
    \centering
    \resizebox{0.99\textwidth}{!}{%
    \begin{tabular}{ccc}
        \begin{subfigure}[t]{0.32\textwidth}
            \centering
            \includegraphics[width=\textwidth]{figure/_1_last_token/SST2_LastToken_1B.pdf}
            \caption{1B on SST-2}
        \end{subfigure} &
        \begin{subfigure}[t]{0.32\textwidth}
            \centering
            \includegraphics[width=\textwidth]{figure/_1_last_token/SST2_LastToken_3B.pdf}
            \caption{3B on SST-2}
        \end{subfigure} &
        \begin{subfigure}[t]{0.32\textwidth}
            \centering
            \includegraphics[width=\textwidth]{figure/_1_last_token/SST2_LastToken_8B.pdf}
            \caption{8B on SST-2}
        \end{subfigure} \\

        \begin{subfigure}[t]{0.32\textwidth}
            \centering
            \includegraphics[width=\textwidth]{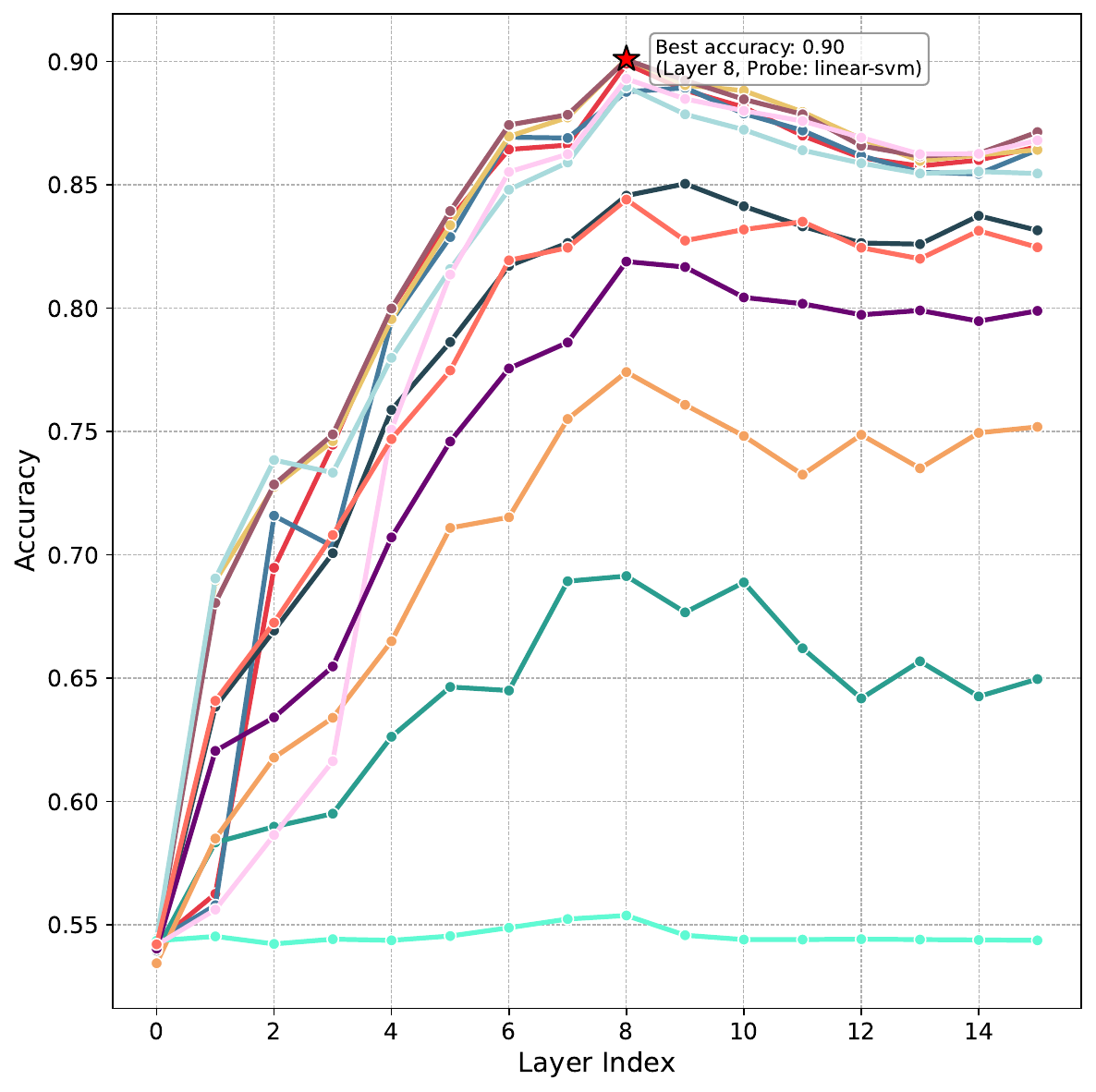}
            \caption{1B on IMDB}
        \end{subfigure} &
        \begin{subfigure}[t]{0.32\textwidth}
            \centering
            \includegraphics[width=\textwidth]{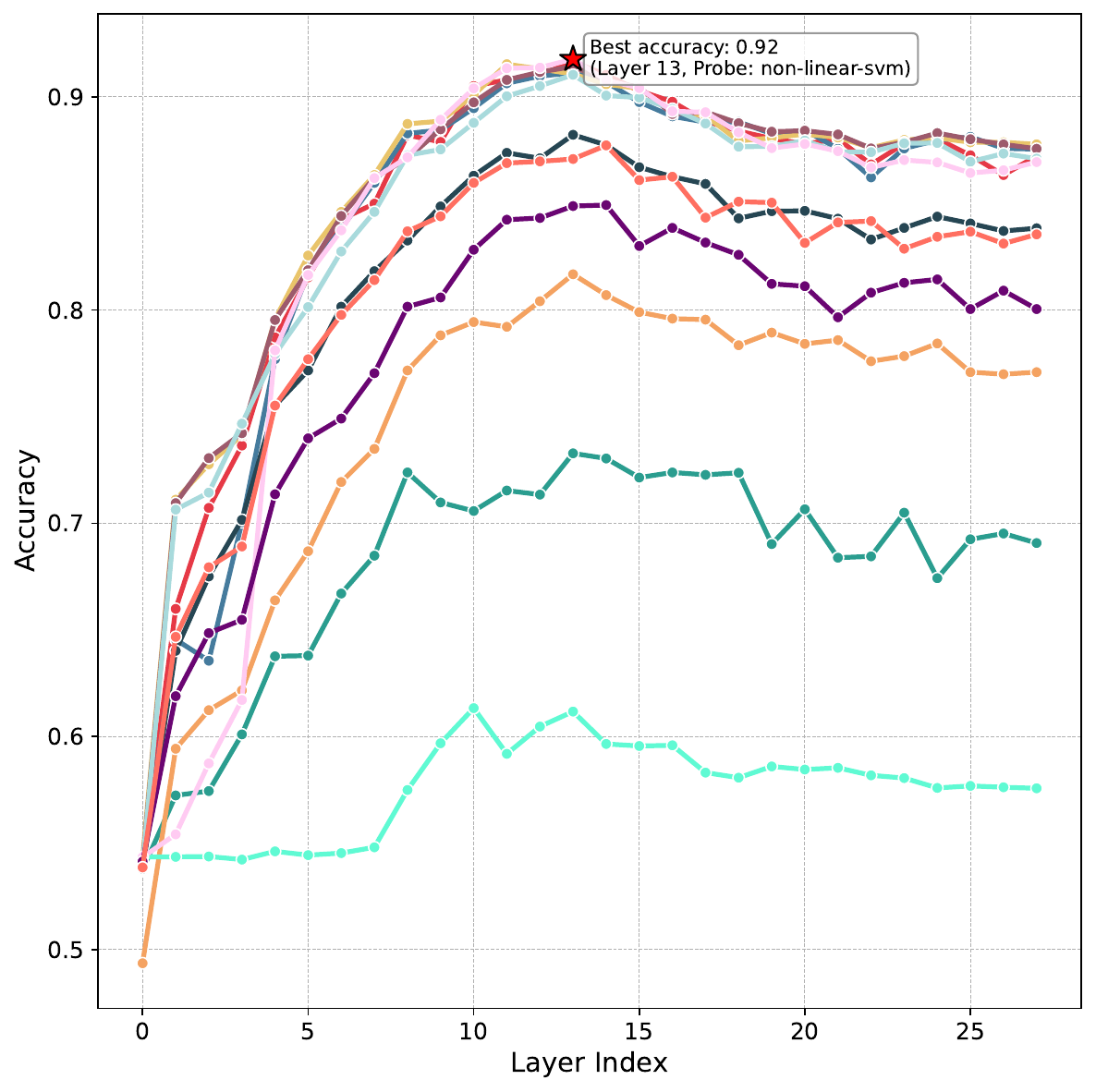}
            \caption{3B on IMDB}
        \end{subfigure} &
        \begin{subfigure}[t]{0.32\textwidth}
            \centering
            \includegraphics[width=\textwidth]{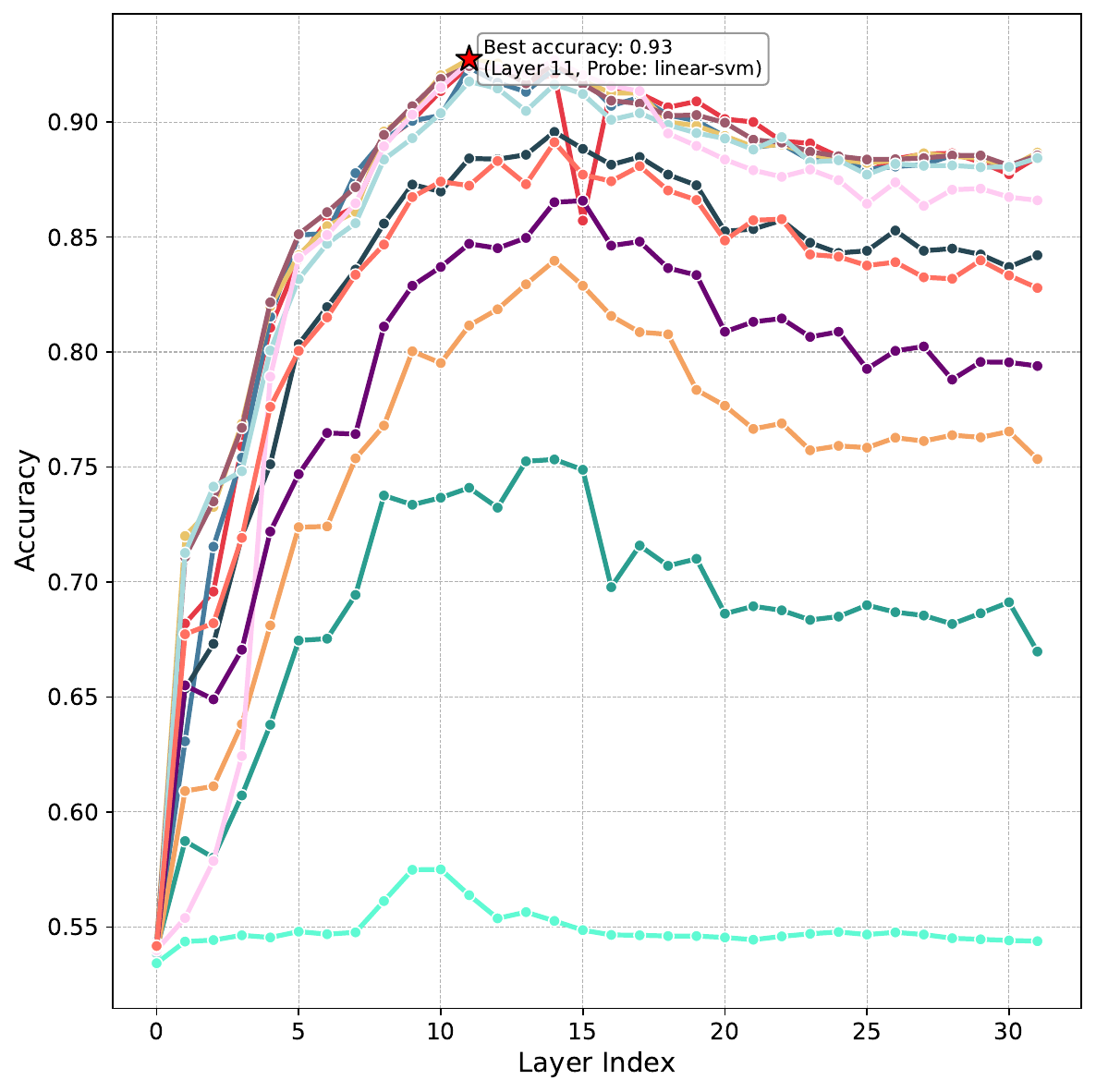}
            \caption{8B on IMDB}
        \end{subfigure} \\

        \begin{subfigure}[t]{0.32\textwidth}
            \centering
            \includegraphics[width=\textwidth]{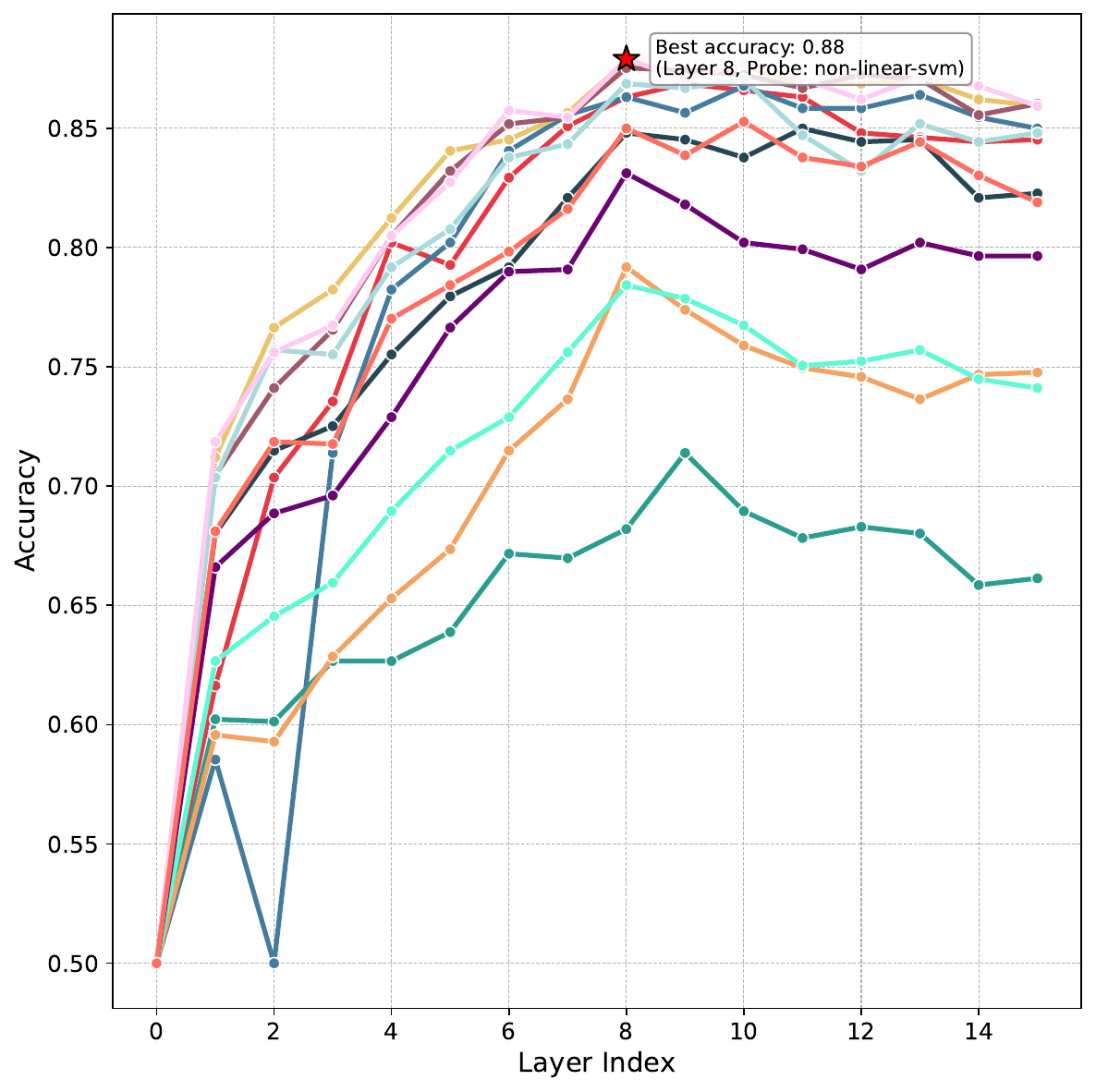}
            \caption{1B on Rotten Tomatoes}
        \end{subfigure} &
        \begin{subfigure}[t]{0.32\textwidth}
            \centering
            \includegraphics[width=\textwidth]{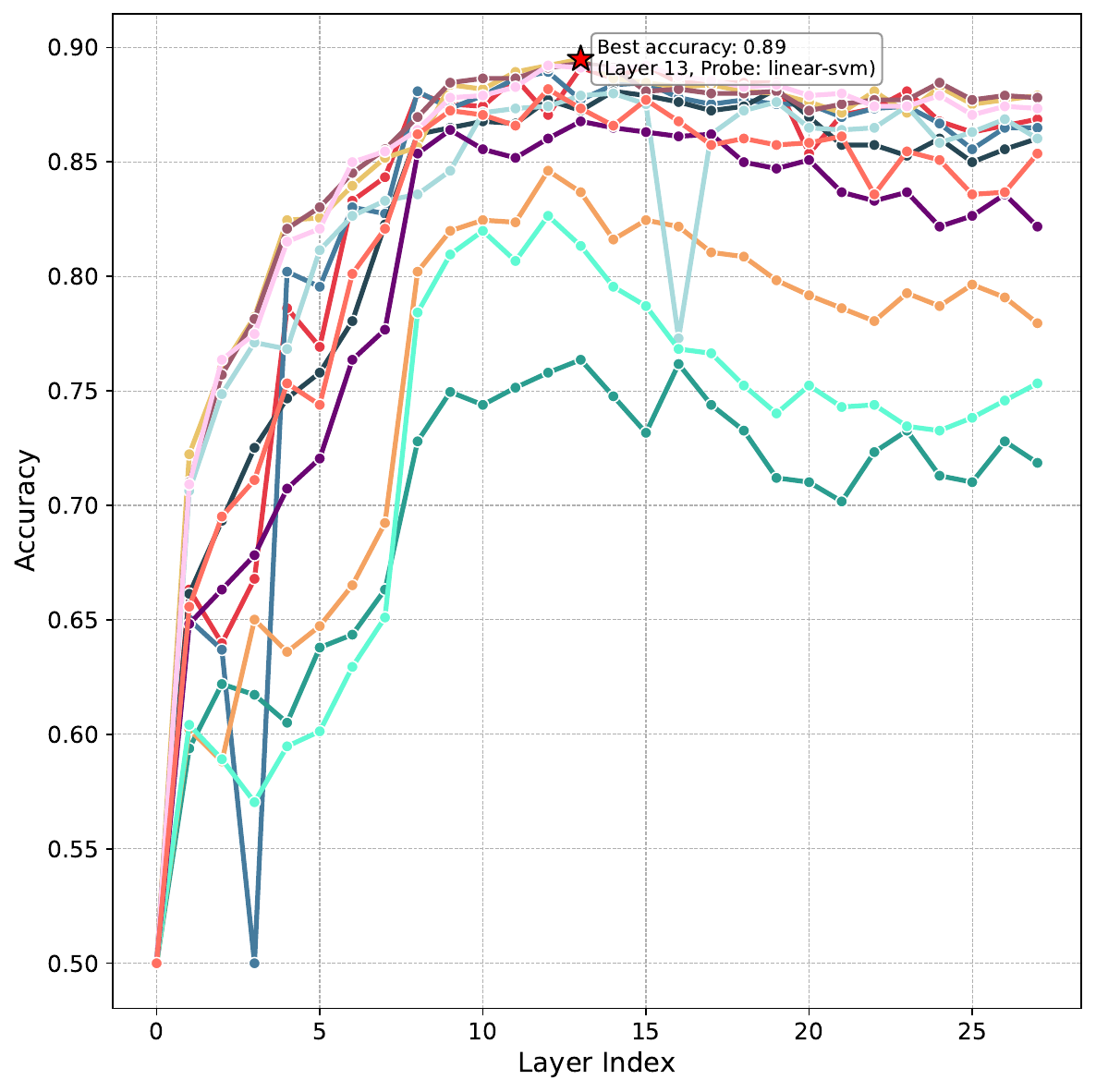}
            \caption{3B on Rotten Tomatoes}
        \end{subfigure} &
        \begin{subfigure}[t]{0.32\textwidth}
            \centering
            \includegraphics[width=\textwidth]{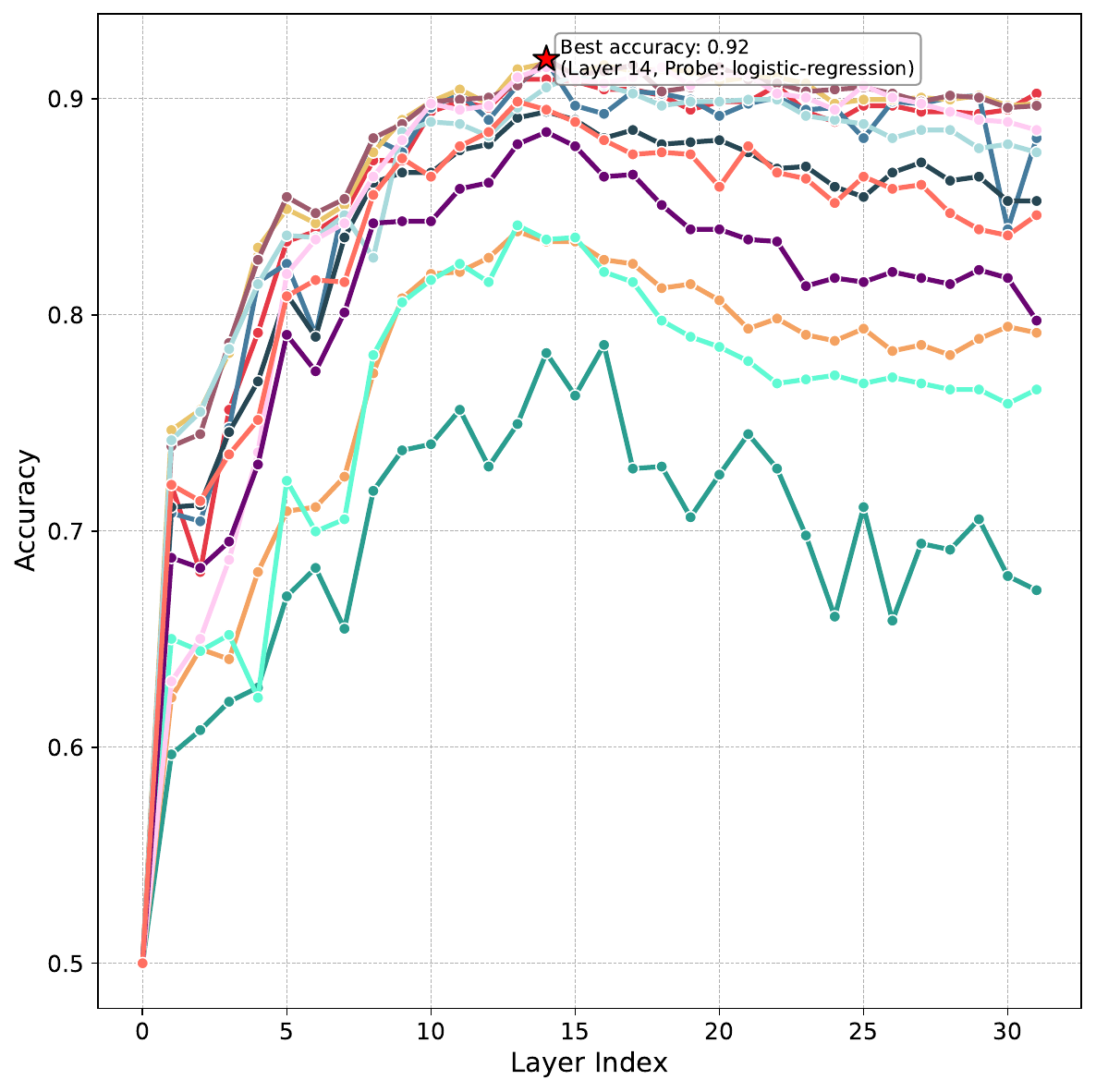}
            \caption{8B on Rotten Tomatoes}
        \end{subfigure} \\

        \begin{subfigure}[t]{0.32\textwidth}
            \centering
            \includegraphics[width=\textwidth]{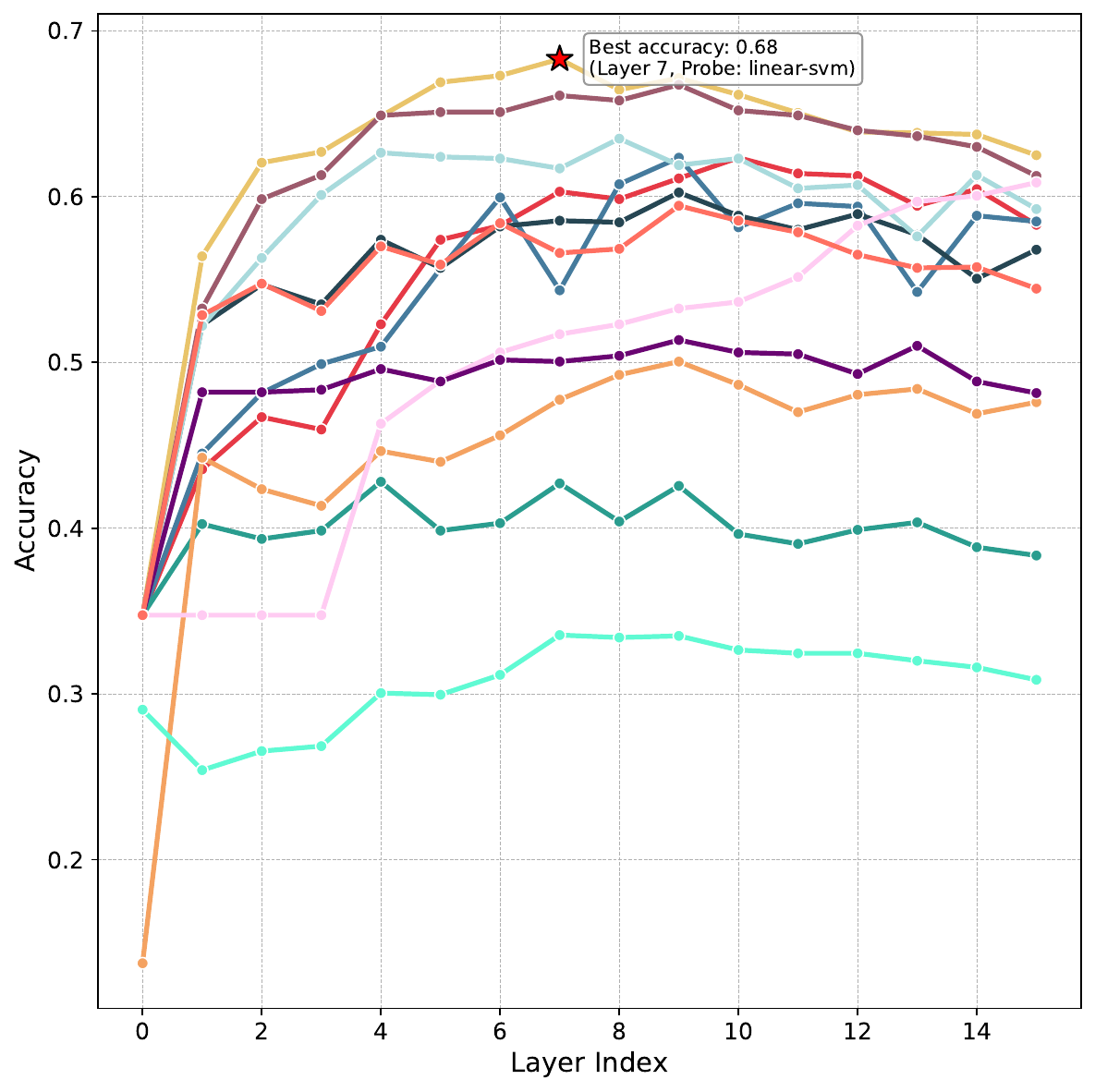}
            \caption{1B on Emotion}
        \end{subfigure} &
        \begin{subfigure}[t]{0.32\textwidth}
            \centering
            \includegraphics[width=\textwidth]{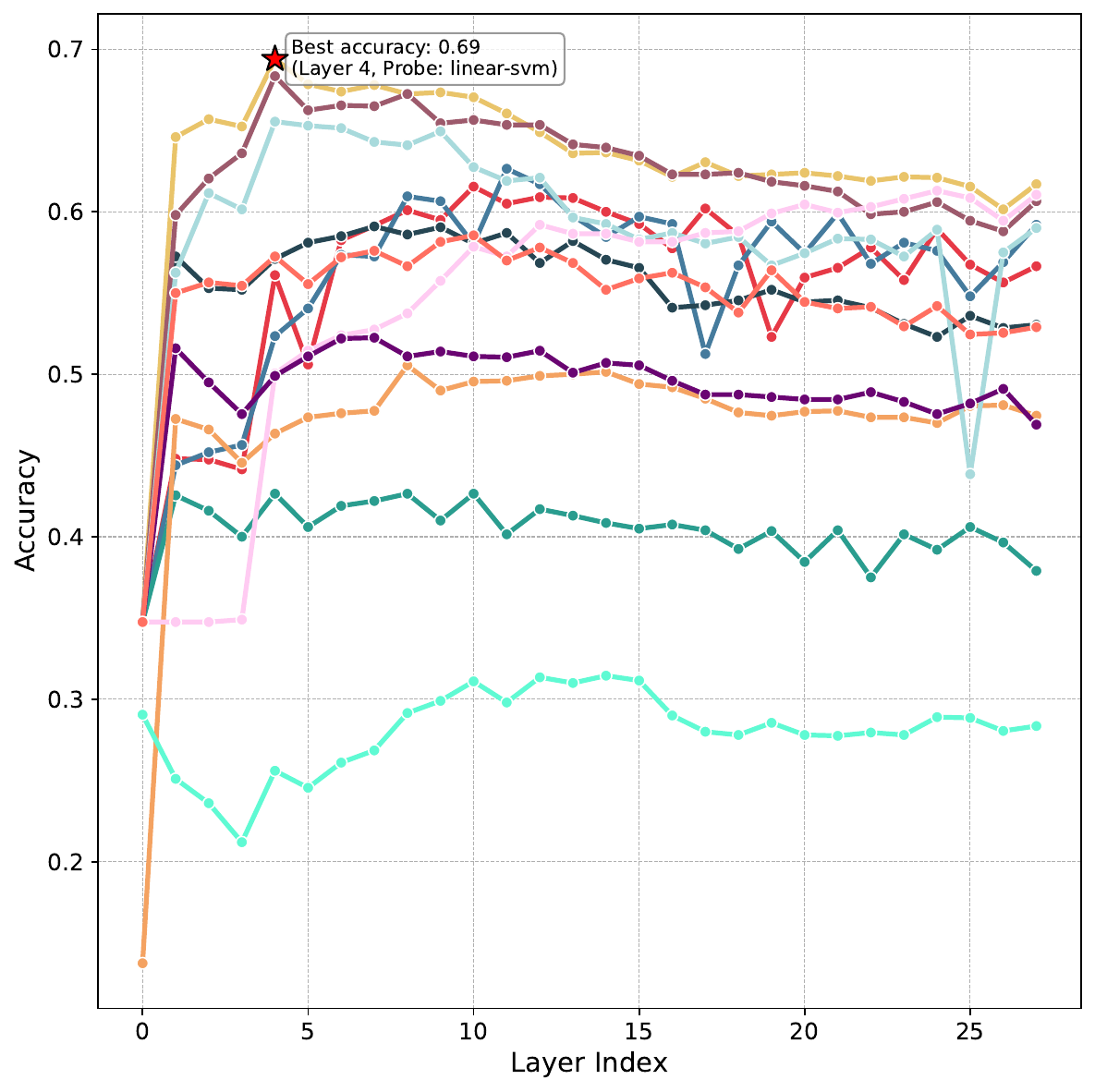}
            \caption{3B on Emotion}
        \end{subfigure} &
        \begin{subfigure}[t]{0.32\textwidth}
            \centering
            \includegraphics[width=\textwidth]{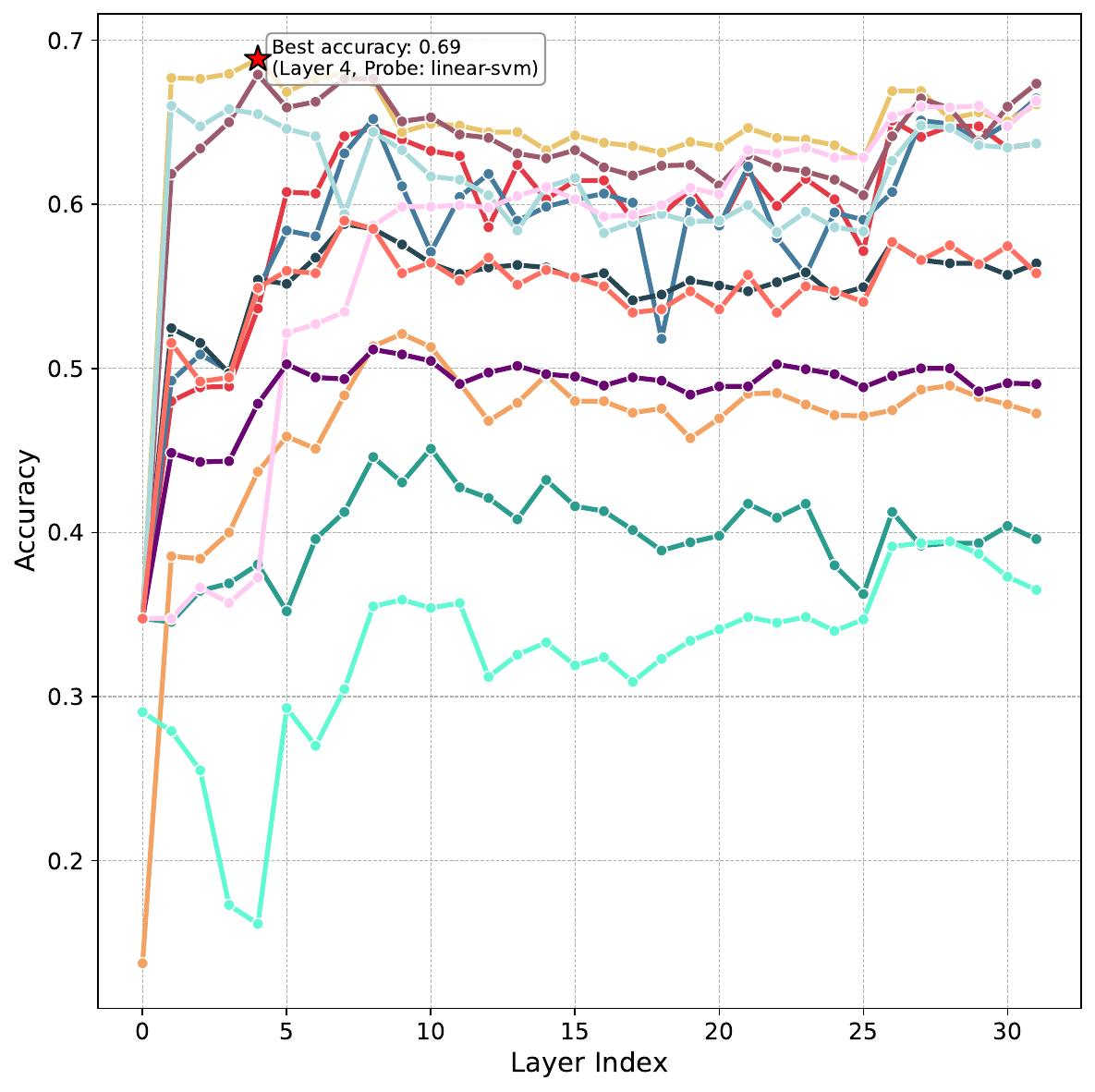}
            \caption{8B on Emotion}
        \end{subfigure} \\
    \end{tabular}
    }
    \begin{center}
        \resizebox{\textwidth}{!}{ 
            \begin{tikzpicture}
                \begin{axis}[
                    hide axis, 
                    xmin=0, xmax=1, 
                    ymin=0, ymax=1,
                    legend columns=6, 
                    legend style={
                        draw=none, 
                        column sep=1.5ex, 
                        font=\small 
                    },
                    legend entries={
                        bilstm,
                        cnn,
                        decision-tree,
                        knn,
                        lightgbm,
                        linear-svm,
                        logistic-regression,
                        mlp,
                        naive-bayes-gaussian,
                        non-linear-svm,
                        random-forest,
                        xgboost
                    }
                ]
                    \addlegendimage{mark=*, color=bilstm};
                    \addlegendimage{mark=*, color=cnn};
                    \addlegendimage{mark=*, color=decisiontree};
                    \addlegendimage{mark=*, color=knn};
                    \addlegendimage{mark=*, color=lightgbm};
                    \addlegendimage{mark=*, color=linearsvm};
                    \addlegendimage{mark=*, color=logisticregression};
                    \addlegendimage{mark=*, color=mlp};
                    \addlegendimage{mark=*, color=naivebayesgaussian};
                    \addlegendimage{mark=*, color=nonlinearsvm};
                    \addlegendimage{mark=*, color=randomforest};
                    \addlegendimage{mark=*, color=xgboost};
                \end{axis}
            \end{tikzpicture}
        }
    \end{center}
    \caption{Layer-wise probing accuracy using the last token on SST-2, IMDB, Rotten Tomatoes, and Emotion.}
    \label{fig:Full_LastToken_Grid}
\end{figure*}

\section{Layer-wise pooling methods confidence accuracy on SST-2, IMDB, Rotten Tomatoes, and Emotion datasets.} \label{sec:appendix_Confidence_Plot}~
\begin{figure*}[!htbp]
    \centering
    \resizebox{0.99\textwidth}{!}{%
    \begin{tabular}{ccc}
        \begin{subfigure}[t]{0.32\textwidth}
            \centering
            \includegraphics[width=\textwidth]{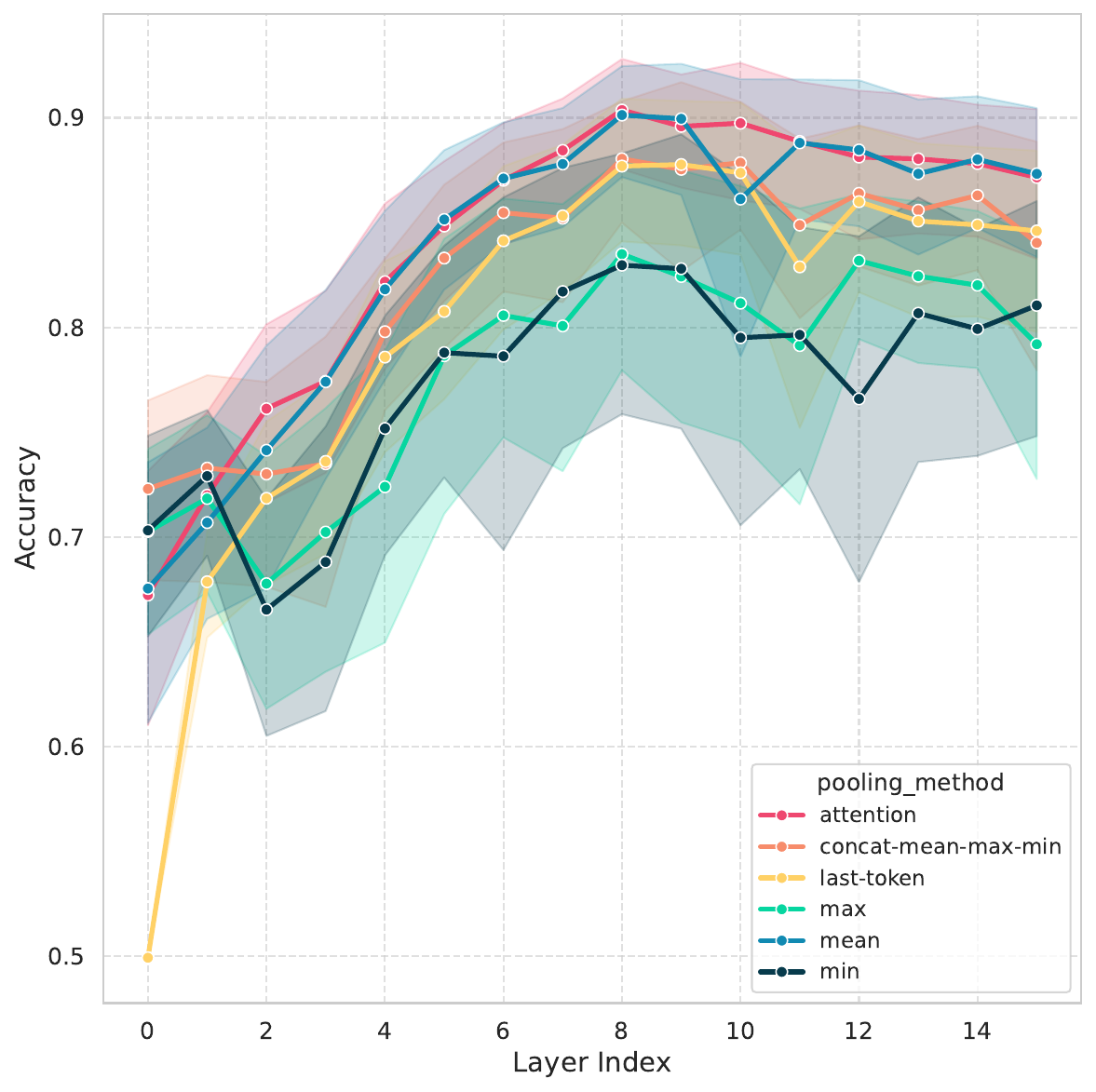}
            \caption{1B-Instruct on SST-2}
        \end{subfigure} &
        \begin{subfigure}[t]{0.32\textwidth}
            \centering
            \includegraphics[width=\textwidth]{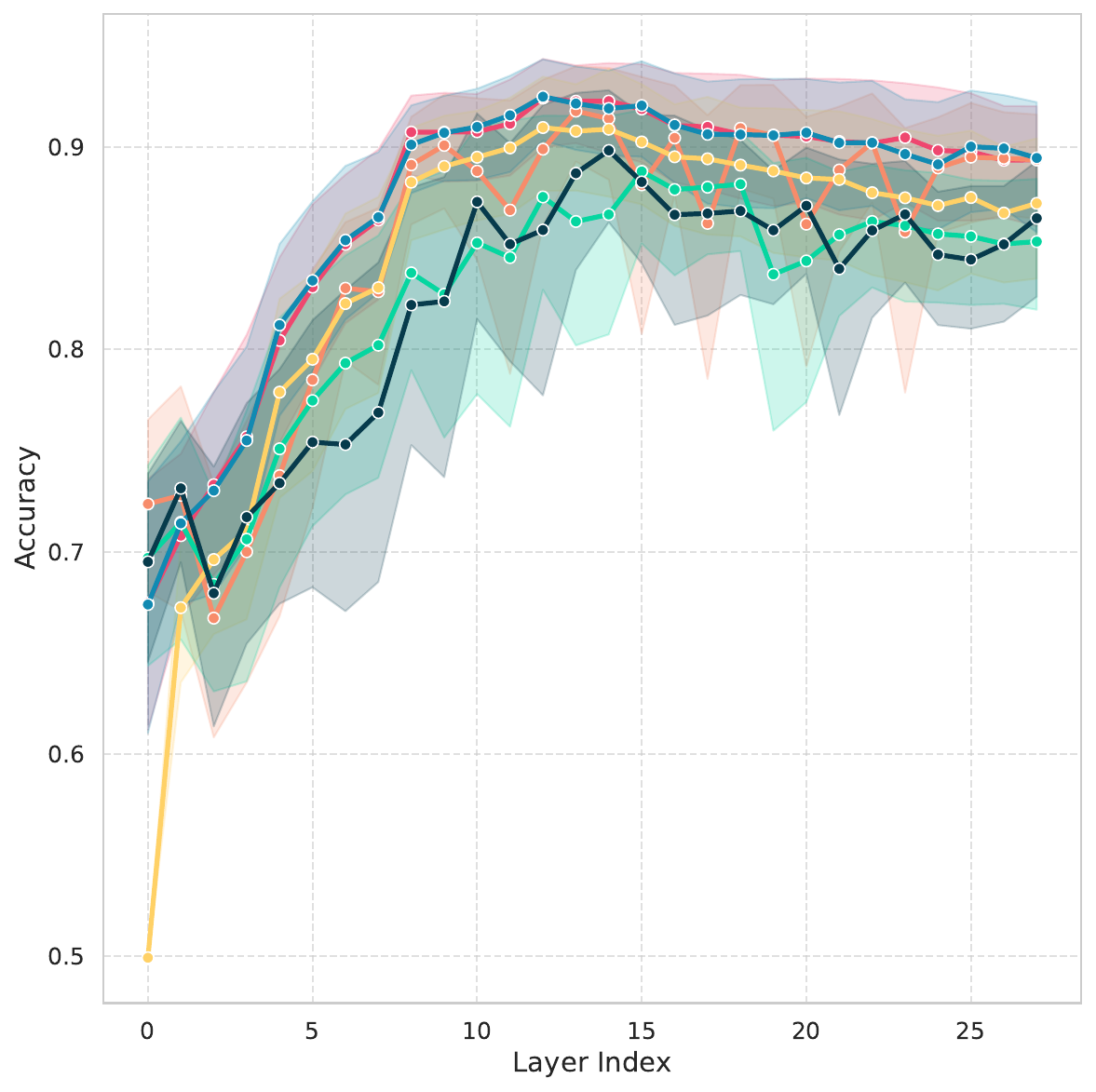}
            \caption{3B-Instruct on SST-2}
        \end{subfigure} &
        \begin{subfigure}[t]{0.32\textwidth}
            \centering
            \includegraphics[width=\textwidth]{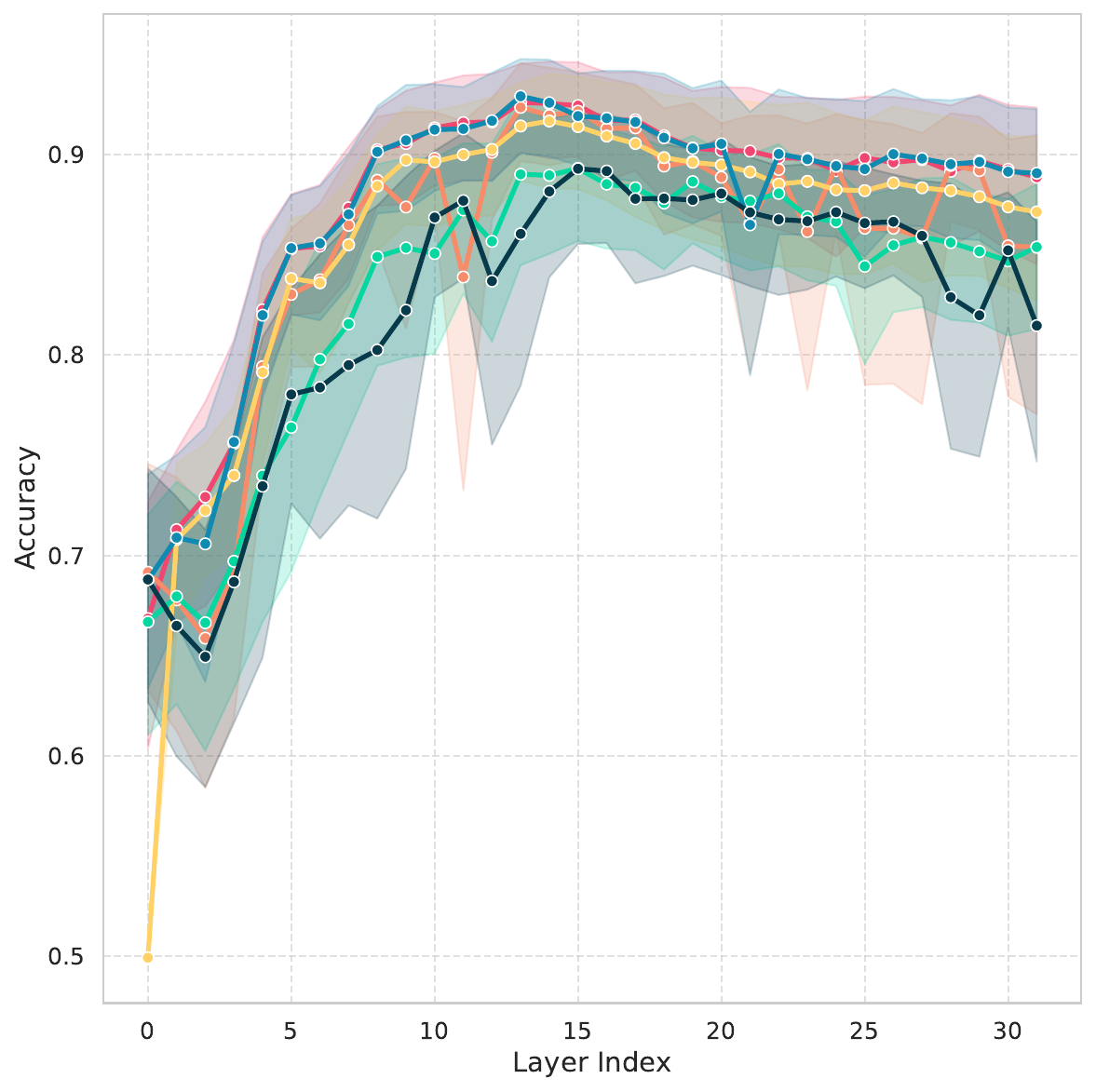}
            \caption{8B-Instruct on SST-2}
        \end{subfigure} \\
        
        \begin{subfigure}[t]{0.32\textwidth}
            \centering
            \includegraphics[width=\textwidth]{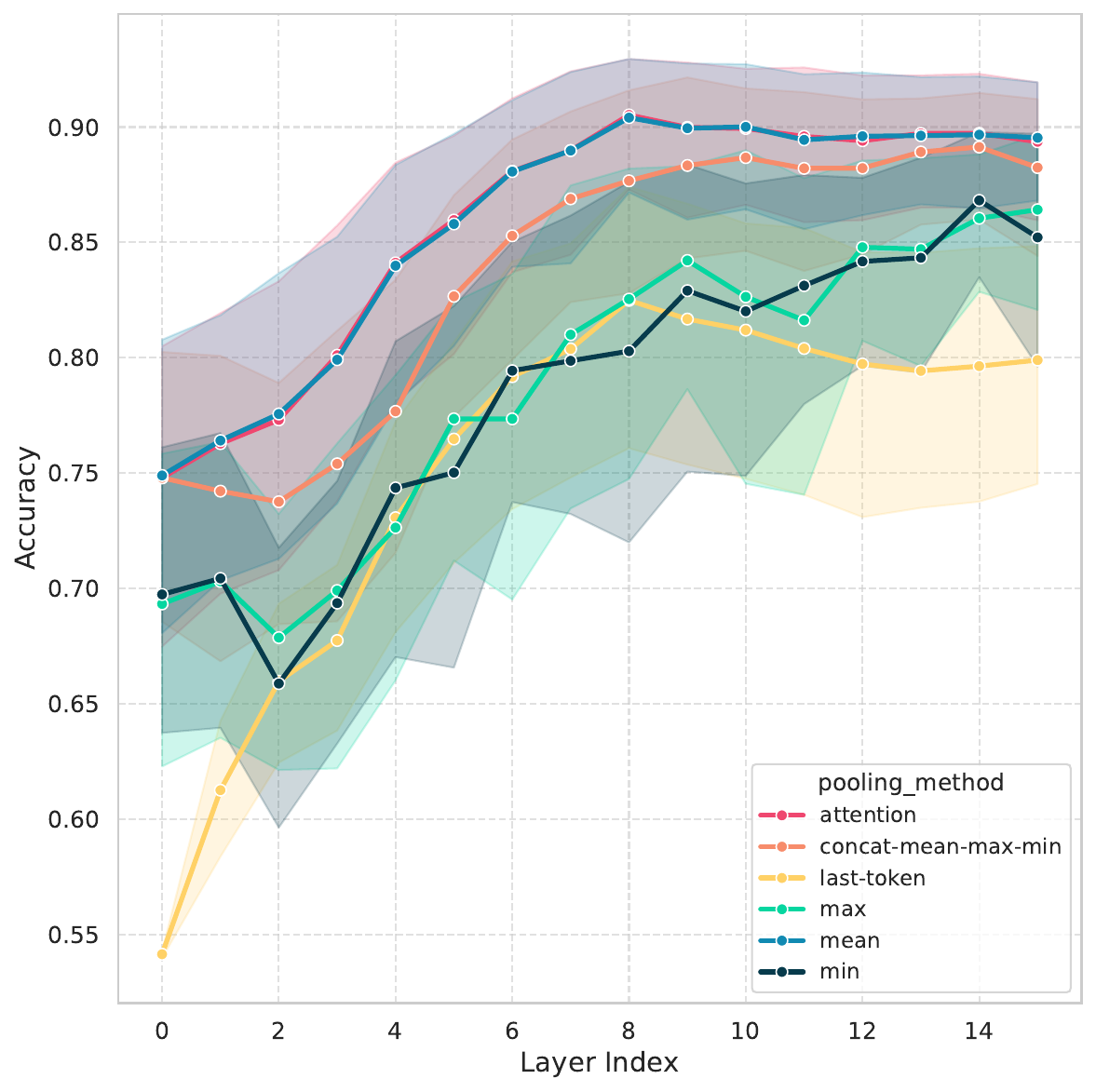}
            \caption{1B-Instruct on IMDB}
        \end{subfigure} &
        \begin{subfigure}[t]{0.32\textwidth}
            \centering
            \includegraphics[width=\textwidth]{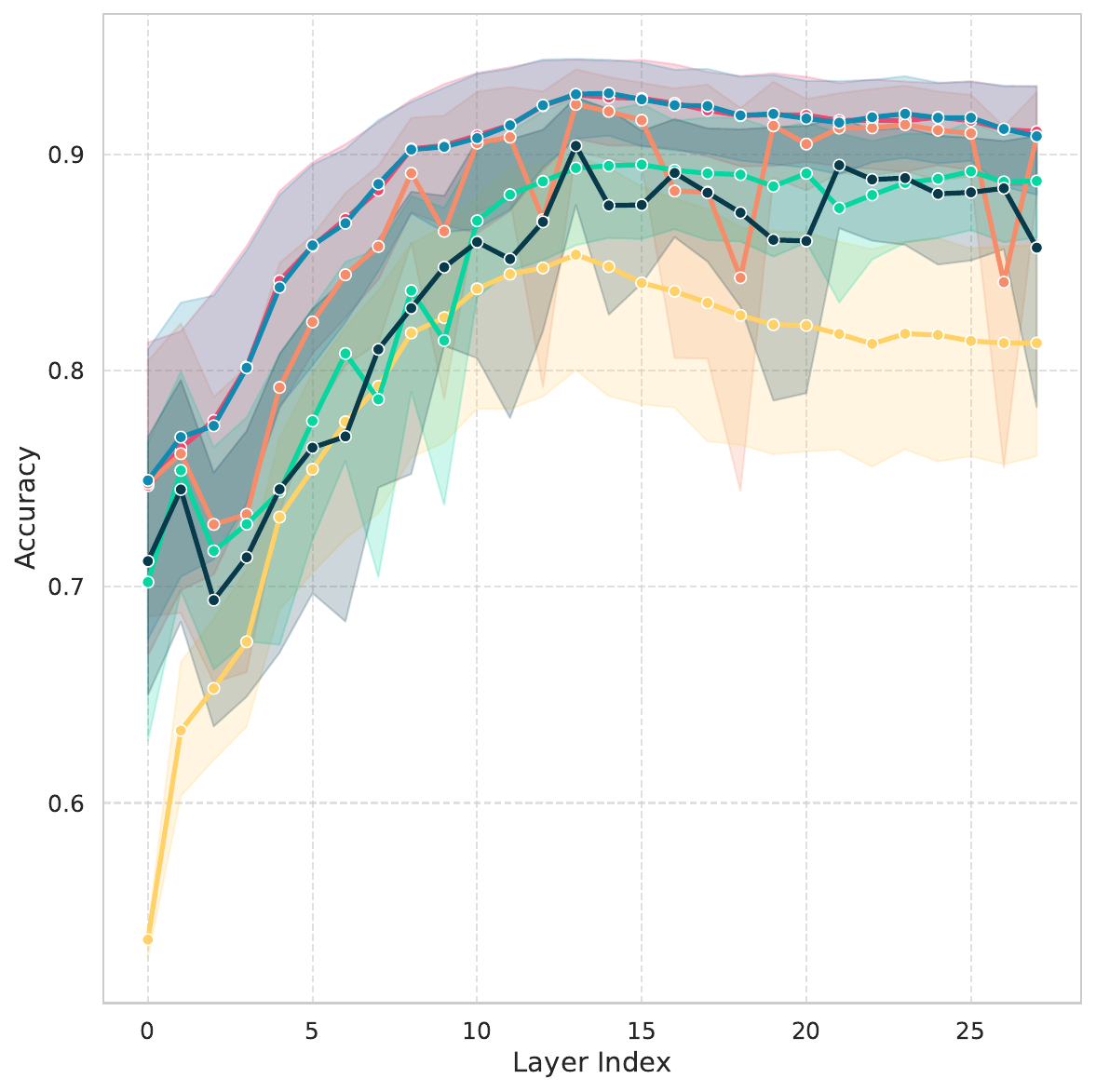}
            \caption{3B-Instruct on IMDB}
        \end{subfigure} &
        \begin{subfigure}[t]{0.32\textwidth}
            \centering
            \includegraphics[width=\textwidth]{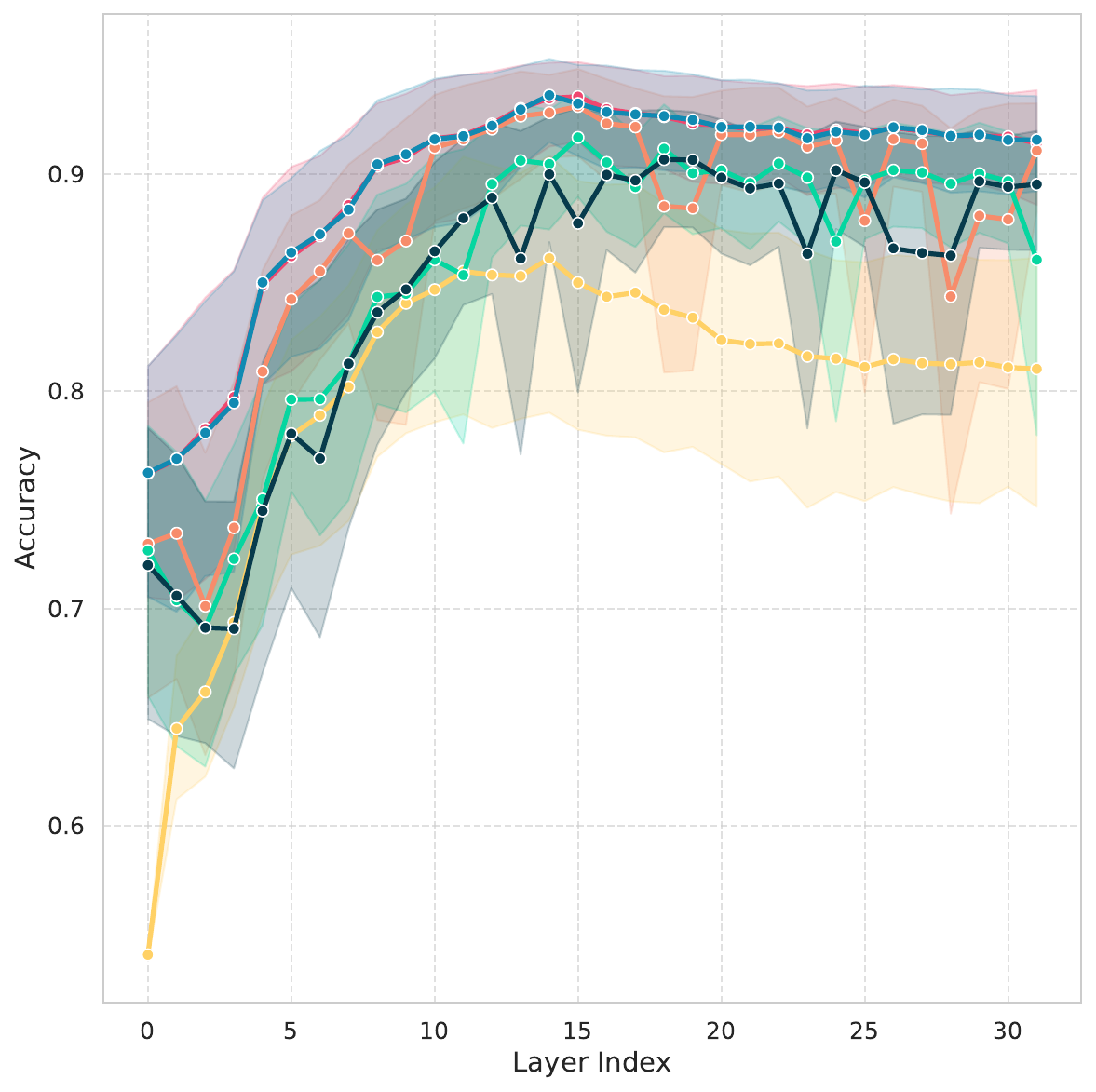}
            \caption{8B-Instruct on IMDB}
        \end{subfigure} \\
        
        \begin{subfigure}[t]{0.32\textwidth}
            \centering
            \includegraphics[width=\textwidth]{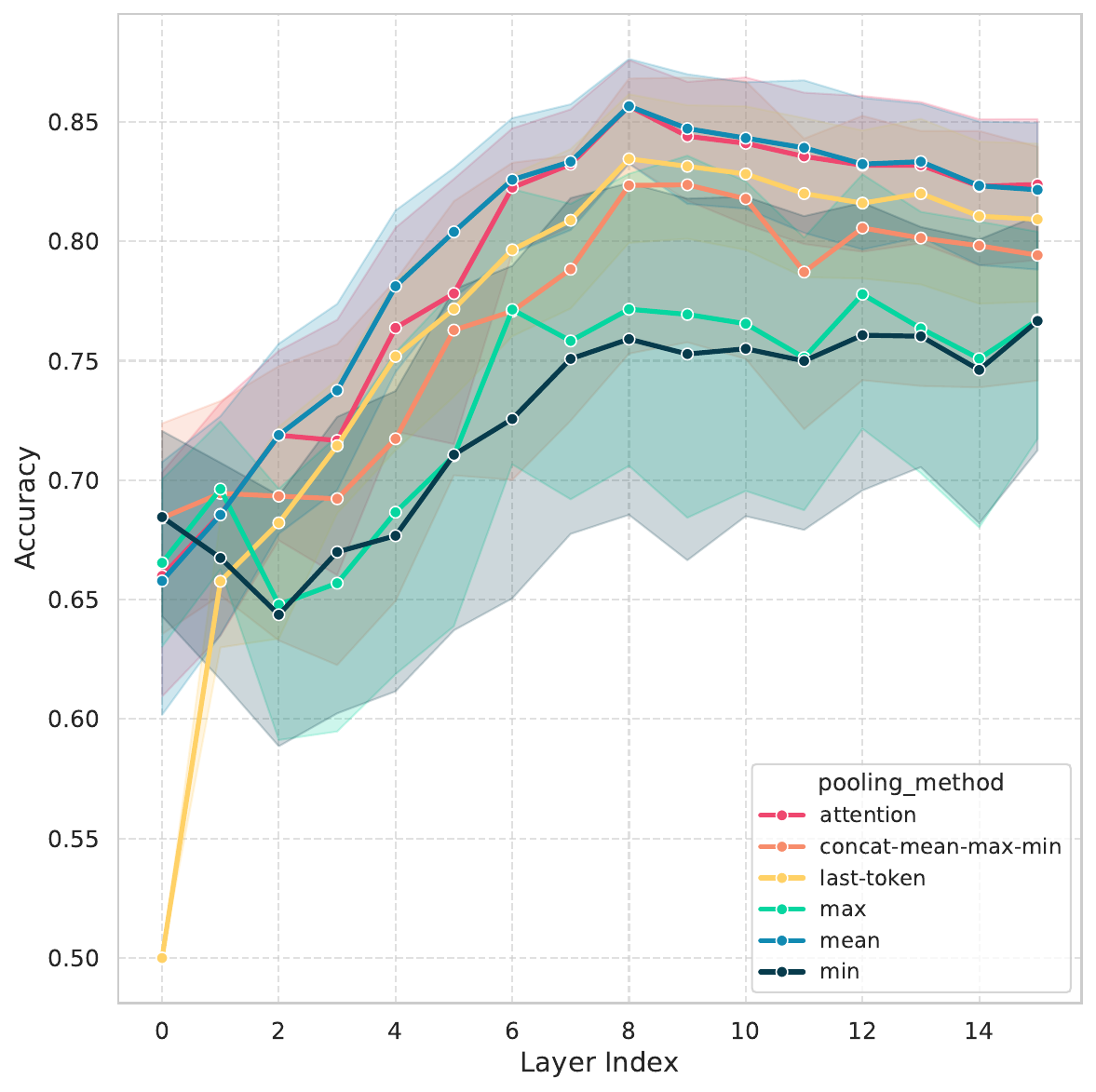}
            \caption{1B-Instruct on Rotten Rotten}
        \end{subfigure} &
        \begin{subfigure}[t]{0.32\textwidth}
            \centering
            \includegraphics[width=\textwidth]{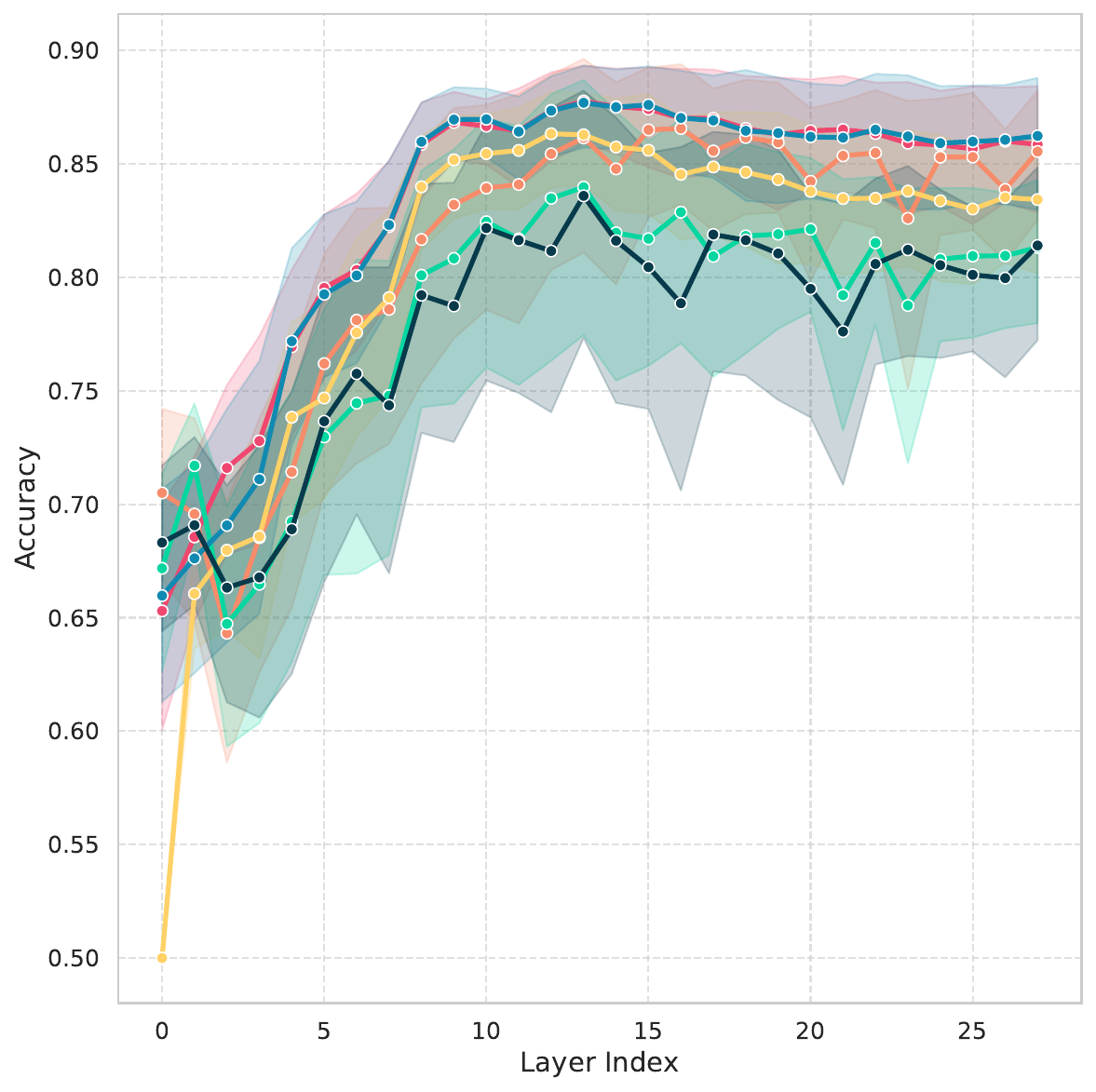}
            \caption{3B-Instruct on Rotten Tomatoes}
        \end{subfigure} &
        \begin{subfigure}[t]{0.32\textwidth}
            \centering
            \includegraphics[width=\textwidth]{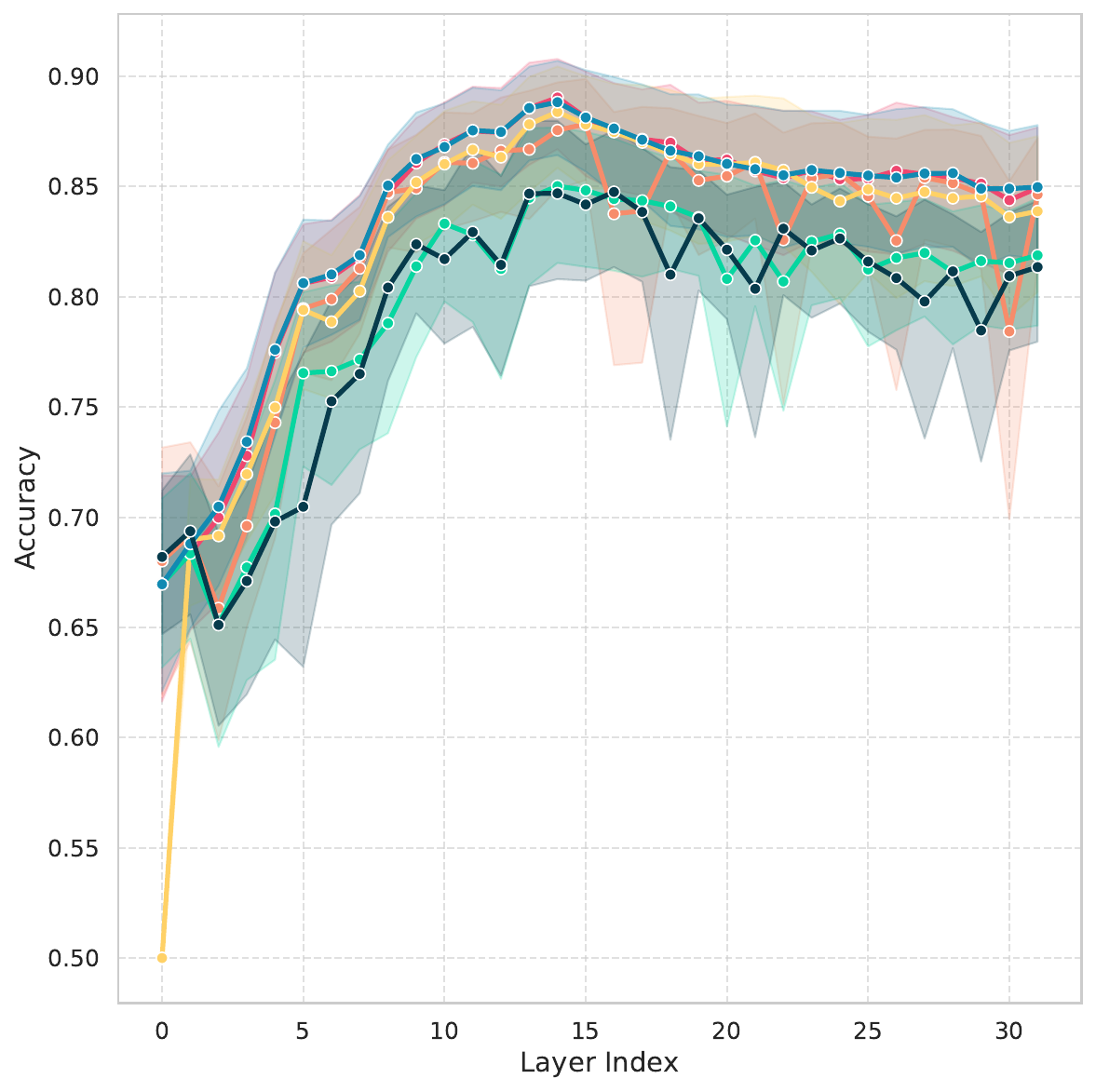}
            \caption{8B-Instruct on Rotten Tomatoes}
        \end{subfigure} \\
    
        \begin{subfigure}[t]{0.32\textwidth}
            \centering
            \includegraphics[width=\textwidth]{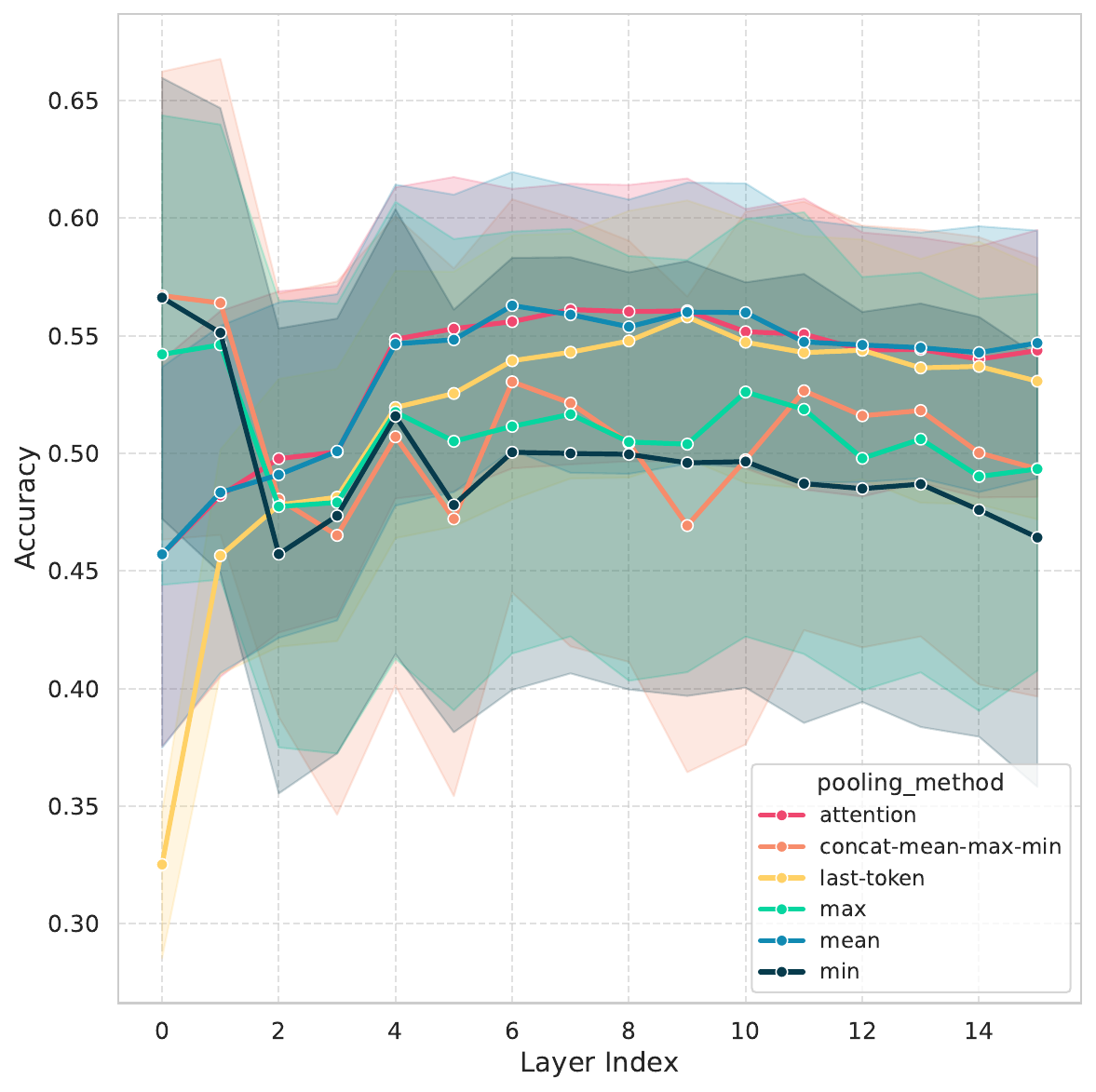}
            \caption{1B-Instruct on Emotion}
        \end{subfigure} &
        \begin{subfigure}[t]{0.32\textwidth}
            \centering
            \includegraphics[width=\textwidth]{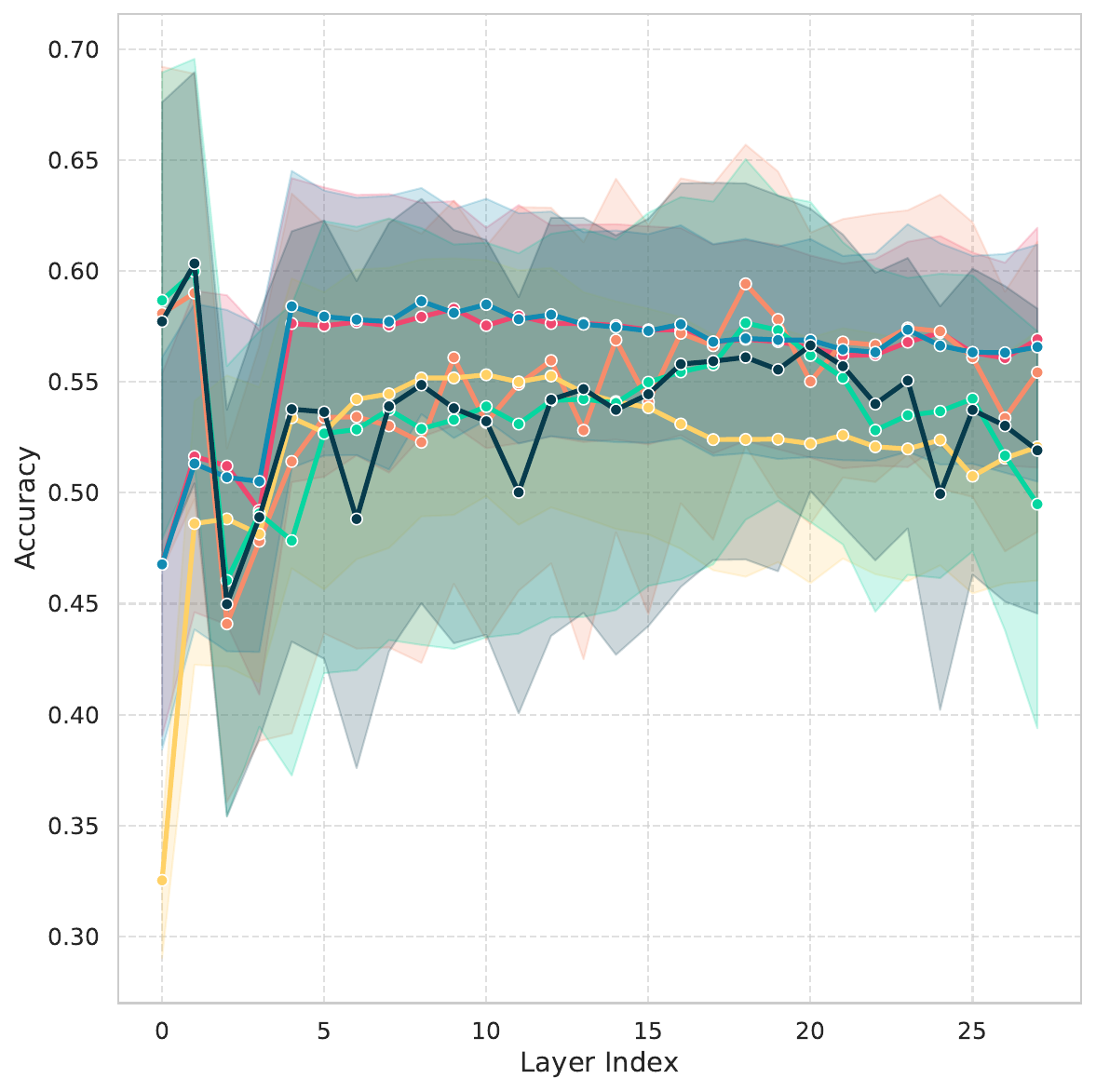}
            \caption{3B-Instruct on Emotion}
        \end{subfigure} &
        \begin{subfigure}[t]{0.32\textwidth}
            \centering
            \includegraphics[width=\textwidth]{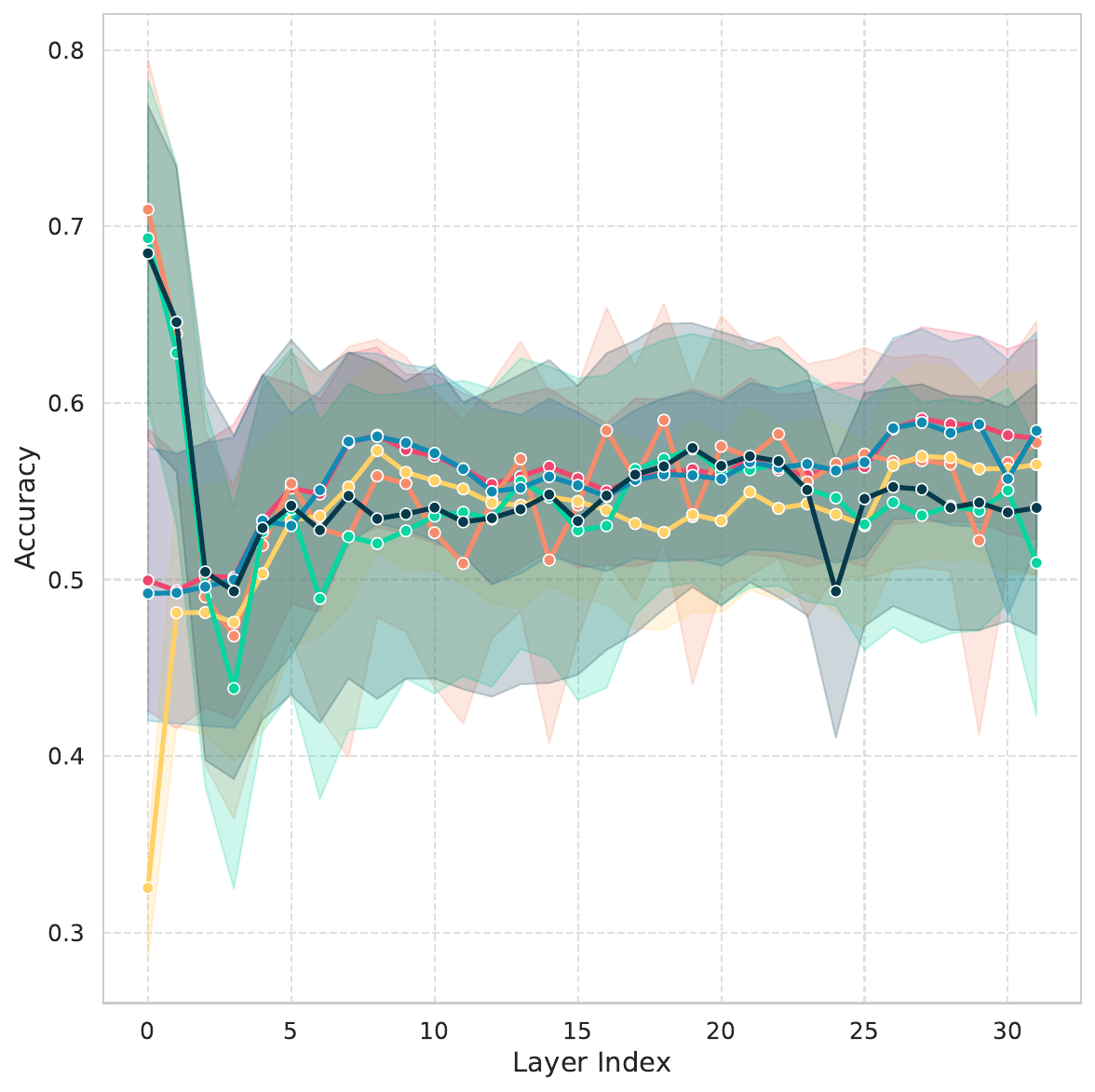}
            \caption{8B-Instruct on Emotion}
        \end{subfigure} \\
    \end{tabular}
    }
    
    \caption{Layer-wise pooling methods confidence accuracy on SST-2, IMDB, Rotten Tomatoes, and Emotion.}
    \label{fig:all_probing}
\end{figure*}

\twocolumn
\section{Compuattional Efficiency of \textsc{SentriLlama} in SST-2, IMDB, Rotten Tomatoes, and Emotion datasets.}\label{sec:appendix_Comp_Efficiency}
\begin{table}[!h]

\resizebox{\columnwidth}{!}{%
\begin{tabular}{@{}lccc@{}}
\toprule
\multirow{2}{*}{\textbf{Model}} & \textbf{Peak GPU} & \textbf{Avg. Time} & \textbf{Throughput} \\ 
& \textbf{Usage} & \textbf{per Sample} & (\textbf{Samples/sec}) \\
\midrule
\textbf{Instruct-Llama 3.2 (1B)} & 4.8GB & 11.33 ms & 88 \\
\textbf{Instruct-Llama 3.2 (3B)} & 9 GB & 20.83 ms & 48 \\
\textbf{Instruct-Llama 3.1 (8B)} & 18.6 GB & 35.33 ms & 28 \\
\midrule
\textbf{\textsc{SentriLlama} 3.2 (1B)} & 3.8 GB & 7.13 ms & 140 \\ 
\textbf{\textsc{SentriLlama} 3.2 (1B) Instruct} & 3.7 GB & 6.56 ms & 152 \\
\textbf{\textsc{SentriLlama} 3.2 (3B) Instruct} & 3.8 GB & 4.73 ms & 211 \\ 
\textbf{\textsc{SentriLlama} 3.1 (8B) Instruct} & 3.6GB & 6.31 ms & 158 \\ 
\midrule
\textbf{DeBERTa V3 Large (418M) } & 2.7 GB & 22.79 ms & 44 \\
\textbf{RoBERTa Large (355M)} & 2.6 GB & 8.90 ms & 119 \\ 
\bottomrule
\end{tabular}
}
\caption{Computational efficiency comparison of \textsc{SentriLlama}, DeBERTa, RoBERTa, and a prompt-based method on the IMDB dataset.}
\label{tab:Computational_Efficiency_IMDB}
\end{table}
\begin{table}[!h]
\centering

\resizebox{\columnwidth}{!}{%
\begin{tabular}{@{}lccc@{}}
\toprule
\multirow{2}{*}{\textbf{Model}} & \textbf{Peak GPU} & \textbf{Avg. Time} & \textbf{Throughput} \\ 
& \textbf{Usage} & \textbf{per Sample} & (\textbf{Samples/sec}) \\
\midrule
\textbf{Instruct-Llama 3.2 (1B)} & 2.4 GB & 12.01 ms & 83 \\
\textbf{Instruct-Llama 3.2 (3B)} & 6.2 GB & 18.41 ms & 54 \\
\textbf{Instruct-Llama 3.1 (8B)} & 15.4 GB & 21.17 ms & 50 \\
\midrule
\textbf{\textsc{SentriLlama} 3.2 (1B)} & 1.6 GB & 7.08 ms & 141 \\ 
\textbf{\textsc{SentriLlama} 3.2 (1B) Instruct} & 1.5 GB & 6.42 ms & 156 \\
\textbf{\textsc{SentriLlama} 3.2 (3B) Instruct} & 1.4 GB & 4.80 ms & 208 \\ 
\textbf{\textsc{SentriLlama} 3.1 (8B) Instruct} & 5.7GB & 19.30 ms & 97 \\ 
\midrule
\textbf{DeBERTa V3 Large (418M) } & 845 MB & 23.32 ms & 43 \\
\textbf{RoBERTa Large (355M)} & 692 MB & 9.05 ms & 118 \\ 
\bottomrule
\end{tabular}
}
\caption{Computational efficiency comparison of \textsc{SentriLlama}, DeBERTa, RoBERTa, and a prompt-based method on the Rotten Tomatoes dataset.}
\label{tab:Computational_Efficiency_Rotten}
\end{table}
\begin{table}[!h]
\centering

\resizebox{\columnwidth}{!}{%
\begin{tabular}{@{}lccc@{}}
\toprule
\multirow{2}{*}{\textbf{Model}} & \textbf{Peak GPU} & \textbf{Avg. Time} & \textbf{Throughput} \\ 
& \textbf{Usage} & \textbf{per Sample} & (\textbf{Samples/sec}) \\
\midrule
\textbf{Instruct-Llama 3.2 (1B)} & 2.4 GB & 10.97 ms & 91 \\
\textbf{Instruct-Llama 3.2 (3B)} &  6.2 GB & 17.42 ms & 57\\
\textbf{Instruct-Llama 3.1 (8B)} & 15.4 GB & 20.53 ms & 49 \\
\midrule
\textbf{\textsc{SentriLlama} 3.2 (1B)} & 673 MB & 2.70 ms & 370 \\ 
\textbf{\textsc{SentriLlama} 3.2 (1B) Instruct} & 673 MB & 2.07 ms & 483 \\
\textbf{\textsc{SentriLlama} 3.2 (3B) Instruct} & 1 GB & 2.76 ms & 363 \\ 
\textbf{\textsc{SentriLlama} 3.1 (8B) Instruct} & 2.6GB & 3.84 ms & 260 \\ 
\midrule
\textbf{DeBERTa V3 Large (418M) } & 844 MB & 22.52 ms & 44 \\
\textbf{RoBERTa Large (355M)} & 692 MB & 8.50 ms & 120 \\ 
\bottomrule
\end{tabular}
}
\caption{Computational efficiency comparison of \textsc{SentriLlama}, DeBERTa, RoBERTa, and a prompt-based method on the Emotion dataset.}
\label{tab:Computational_Efficiency_Emotion}
\end{table}
\end{document}